\documentclass[]{jair}

\setcopyright{cc}
\copyrightyear{2026}
\acmDOI{10.1613/jair.1.22089} 

\JAIRAE{Tongliang Liu}
\JAIRTrack{} 
\acmVolume{87}  
\acmArticle{19} 
\acmMonth{11}   
\acmYear{2026}  

\RequirePackage[
  datamodel=acmdatamodel,
  style=acmauthoryear,
  backend=biber,
  giveninits=true,
  uniquename=init
  ]{biblatex}

\usepackage{macros}

\usepackage{bm}
\usepackage{multirow}
\usepackage{nicematrix}
\usepackage{makecell}
\usepackage{subcaption}

\usepackage{tikz}
\usepackage{pifont}%

\usepackage{diagbox}
\newcommand{\cmark}{\ding{51}}%
\newcommand{\xmark}{\ding{55}}%

\begin{document}

\title[BCAT]{BCAT: A Block Causal Transformer for PDE Foundation Models for Fluid Dynamics}
\def\model{\textsc{bcat}~}
\author{Yuxuan Liu}
\orcid{0000-0001-5788-1329}
\email{yxliu@ucla.edu}
\affiliation{%
  \institution{University of California, Los Angeles}
  \city{Los Angeles}
  \state{California}
  \country{United States}
}

\author{Jingmin Sun}
\orcid{0000-0002-5728-2606}
\email{jsun143@jh.edu}
\affiliation{%
  \institution{Johns Hopkins University}
  \city{Baltimore}
  \state{Maryland}
  \country{United States}}

\author{Hayden Schaeffer}
\authornote{Corresponding Author.}
\orcid{0000-0003-1379-1238}
\email{hayden@math.ucla.edu}
\affiliation{%
  \institution{University of California, Los Angeles}
  \city{Los Angeles}
  \state{California}
  \country{United States}
}

\renewcommand{\shortauthors}{Liu, Sun \& Schaeffer}

\begin{abstract}
{\bf Background:}
    Foundation models have demonstrated remarkable generalization in vision and language, but their accurate use for scientific computing remains challenging. For two-dimensional fluid-dynamics prediction, current transformer-based models are often hindered by limited context and rigid causality information, as they typically map fixed windows to the next step.
    
    {\bf Objectives:}
    We aim to develop a unified architecture for heterogeneous fluid-dynamics families. The goal is to better capture the complex spatio-temporal dependencies inherent in nonlinear fluid dynamics.
    
    {\bf Methods:}
    We introduce BCAT, a block-causal transformer for autoregressive prediction of two-dimensional fluid-dynamics solutions. Our approach adapts block-causal attention to next-frame prediction, leveraging previous frames as contextual priors. The model is trained on 69K trajectory sequences from PDEBench, PDEArena, and CFDBench using the Muon optimizer.
    
    {\bf Results:}
    BCAT achieved an average relative $L^2$ error of 1.18\% across the six evaluated fluid families, reducing error by 45\% relative to a 3.4-times-larger baseline. In ablations, next-frame prediction was 3.5x more accurate than the corresponding next-token variant. A five-seed analysis, noisy-input stress test, long-rollout analysis, and physics diagnostics further characterize the model.
    
    {\bf Conclusions:}
    The block-causal framework captures spatial interactions within frames while enforcing temporal causality across frames. BCAT provides an accurate and computationally efficient model for the two-dimensional fluid-dynamics benchmarks considered here.
\end{abstract}


\received{27 February 2026}
\received[accepted]{12 October 2026}

\maketitle

\section{Introduction} \label{sec:intro}

Fluid mechanics is a fundamental area of study within physics and engineering, describing a wide range of phenomena through modeling of pressure, velocity, and viscosity. It is used in various applications, including climate forecasting, energy generation in wind and hydropower systems, aerodynamics and aircraft design, blood circulation analysis, and more. 
Due to the highly nonlinear interactions among multiple physical scales, predicting and simulating fluid behavior remains a challenging task. The computational problems become even more difficult in settings with limited observations or noisy measurements. Hence, accurately predicting fluid behavior in such regimes remains a challenge for scientific machine learning. 

While foundation models like GPT \cite{radford2018improving,radford2019language, brown2020language}, DALL-E \cite{ramesh2021zero,ramesh2022hierarchical}, Stable Diffusion \cite{rombach2022high}, LLAMA \cite{touvron2023llama, touvron2023llama2}, Phi \cite{abdin2024phi3, abdin2024phi4}, Gemma \cite{team2024gemma}, and Claude have demonstrated remarkable generalization capabilities in language and vision tasks \cite{bommasani2021opportunities}, they have not accurately addressed scientific computing tasks such as simulating partial differential equations (PDEs). These tasks not only require precise predictions but also the ability to learn the underlying physical rules.
PDE foundation models aim to address this challenge by using a single deep neural network (DNN) to simultaneously encode and approximate the behavior of a variety of physical models. The objective is to develop a unified DNN capable of predicting multiple physical systems with the ability to simulate solutions in unseen parameter regimes, i.e. out-of-domain predictions. Recent progress in PDE foundation models includes the Predicting Operators and Symbolic Expressions (PROSE) \cite{liu2024prose, sun2024towards, sun2024lemon,jollie2024time,liu2024prosefd}, Multiple Physics Pretraining (MPP) \cite{mccabe2023multiple}, In-Context Operator Network (ICON) \cite{yang2023context, yang2023prompting, yang2024pde, cao2024vicon}, DPOT \cite{hao2024dpot}, Fourier Forecasting Network (FourCastNet) \cite{pathak2022fourcastnet}, Poseidon \cite{herde2024poseidon}, and Aurora \cite{bodnar2024aurora}.
~\\\noindent\textbf{Main Contributions:} 
One of the key technical components of foundation models is the transformer architecture, known for its ability to efficiently model complex and long-range relationships in data through attention mechanisms. However, current transformer structures utilized in PDE foundation models are hindered by limited context and causality information, since many existing approaches learn to map a fixed window to the next step. 
In this work, we introduce the \textbf{BCAT} model, based on the \textbf{B}lock \textbf{CA}usal \textbf{T}ransformer architecture\footnote{The code is available at: \url{https://github.com/felix-lyx/bcat}.}, adapting the attention mechanism for spatio-temporal dynamics modeling. The \model model performs next frame prediction instead of next token prediction, which enhances computational efficiency, allows for more token interactions to capture complex dependencies, and utilizes flexible history context windows. Our contributions are listed below.
\begin{itemize}
\item We adapt a block-causal transformer structure to autoregressive prediction of two-dimensional fluid dynamics. BCAT uses next-frame prediction to model spatiotemporal context and achieves a 3.5x accuracy improvement over the corresponding next-token variant.

\item BCAT's next-frame objective enables flexible history windows and parallel training loss computation. At inference, key-value caching further reduces measured wall-clock time for the stated hardware configuration.

\item Across six two-dimensional fluid families from PDEBench, PDEArena, and CFDBench, BCAT obtains the best average relative $L^2$ error in our zero-shot comparison. Relative to the 3.4-times-larger DPOT-L baseline, it reduces the average error by 45\%. We further evaluate low-data adaptation on a turbulence dataset.

\end{itemize}

\section{Related Works} \label{sec:related_works}

\subsection{PDE Foundation Models}
Foundation models are large-scale, pre-trained models that act as flexible bases for a wide range of tasks, leveraging vast amounts of data and computational resources to enable adaptability and generalization across diverse applications \cite{achiam2023gpt,dubey2024llama,yang2024qwen2,bommasani2021opportunities, brooks2024video}. However, their applications to scientific computing are limited, where high precision is needed for numerical tasks. Recently, several approaches have been proposed to explore and develop foundation models for scientific computing. 
PROSE \cite{liu2024prose,sun2024towards,sun2024lemon,liu2024prosefd,jollie2024time} takes multimodal input to construct solution operators, encoding PDEs as symbolic input to provide additional information and allows zero-shot generalization. 
ICON \cite{yang2023context,yang2023prompting,yang2024pde,cao2024vicon} uses an in-context learning approach, allowing the model to perform few-shot learning by observing input-output pairs. 
Some methods focus on the sequential modeling aspect, and various models have been designed to autoregressively predict future states. As additional information is not included, these methods are not able to zero-shot generalize to out-of-distribution data, but can leverage pretraining data to be efficiently fine-tuned \cite{subramanian2024towards}. Examples of such models include MPP \cite{mccabe2023multiple}, which uses axial attention to map a fixed context window to future states, and DPOT \cite{hao2024dpot}, which uses Fourier attention and stabilizes the autoregressive rollout process through a denoising training process.
Poseidon \cite{herde2024poseidon} is a swin-transformer-based model, and the model requires fine-tuning for downstream tasks.

\subsection{Transformers for Time Series}
Time series data takes a similar form as language modeling data, where numerical inputs replace discrete tokens. LLMs and other transformer-based models have been applied to various time series tasks, including forecasting \cite{jin2023time}, classification \cite{liu2021gated}, and anomaly detection \cite{zhou2024can}. One key challenge in adapting pretrained transformers for time series forecasting lies in the encoding of numerical data. In \cite{gruver2024large}, scalar time series are encoded as strings, which can be directly processed by pretrained LLMs. In \cite{chang2024llm4ts}, a separate alignment stage is used to allow LLMs to process patched time series. However, as shown in \cite{tan2024language}, while attention layers can be useful for time series, pretrained LLMs may not outperform simple transformers trained from scratch. Another challenge lies in processing spatio-temporal dynamics, due to the complex interplay between spatial and temporal dependencies. Accurately capturing these interactions often requires models that can handle high-dimensional data, irregular sampling, and non-linear relationships, all while maintaining scalability and computational efficiency. Time-dependent PDEs, climate modeling, and videos can all be seen as examples of spatio-temporal data. The high-dimensional nature requires specific tokenization and processing techniques, such as graph \cite{liu2024can} and Fourier mixer \cite{pathak2022fourcastnet} for spatial dependencies. 
\section{Methods} \label{sec:methods}

\subsection{Problem Setup}
Consider parametric families of two-dimensional time-dependent nonlinear PDEs, whose state-variables of interest are represented by $\u(\x,t) \in \mathbb{R}^d$, where $d$ can be up to 4 in our tests, with $\x\in \Omega\subseteq \R^2$. 
Given data up to $T_0$ frames, i.e. the sequence: \[\{\u(\cdot,t_i)\st 0\le i < T_0\},\] the goal is to predict the subsequent $T$ frames: \[\{\u(\cdot,t_i)\st T_0\le i<T_0+T\}.\] In our experiments, we set $T_0 = T = 10$, that is, 10 frames are given as inputs and the model predicts 10 future frames.

\subsection{Model Overview}
\begin{figure}[t]
\centering
\includegraphics[width=0.6\linewidth]{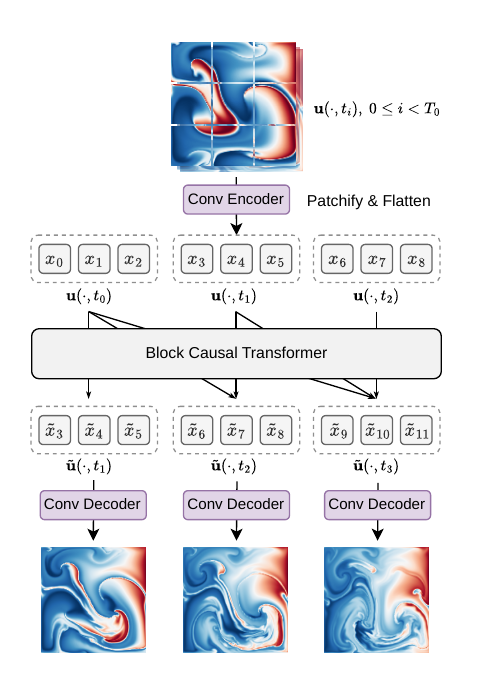}
\caption{\textbf{\model model overview.} The inputs to the model are the initial frames sampled from the datasets, which are patchified and converted into a sequence of features. Transformer layers then take the input sequence to perform next-frame prediction, where a block-causal mask allows spatial interactions within a frame and temporal causality across varying-length context windows. The processed features are projected with a convolutional patch-reconstruction layer to form the final predictions.}\label{fig:model}
\Description{BCAT first maps each input frame to nonoverlapping patch tokens. Decoder-only transformer layers process the temporal token sequence under a block-causal attention mask. A convolutional patch-reconstruction layer maps output tokens back to predicted fluid fields.}
\end{figure}

The \model model is illustrated in Figure \ref{fig:model}. The model comprises two main components: (1) image tokenization and de-tokenization through patch embedding, and (2) the standard decoder-only transformers similar to GPT-2 \cite{radford2019language}. The input data, $\{\u(\cdot,t_i)\st 0\le i<T_0\}$, are first partitioned into patches, which are subsequently flattened and projected into a sequence of visual tokens. The transformer backbone serves as a token mixer, processing these tokens to autoregressively predict future states $\{\u(\cdot,t_i)\st T_0\le i<T_0+T\}$. We use RMS norm \cite{zhang2019root} in place of the usual layer norms and find query-key normalization \cite{henry2020query} helpful in stabilizing the training process.

Training is performed end-to-end by minimizing the mean squared error between the (normalized) predicted and ground truth sequences. Detailed descriptions of the two components are provided in the following sections.

\subsection{Image Tokenization via Patch Embedding}
Inspired by \cite{dosovitskiy2021an}, we employ a patch embedding layer to convert the input data into a sequence of visual tokens. Each frame in the input data, $\u(\cdot,t_i)$, is first divided into non-overlapping patches, which are then flattened and projected into a sequence of visual tokens. For a maximum of $(T_0+T)$ data frames, we simply concatenate the visual tokens from each frame to form the input sequence to the transformer backbone. 

Let $R_x$ and $R_y$ denote the space resolutions and $c$ be the number of channels (physical fields), i.e., $\u(\cdot,t)\in \R^{R_x\times R_y\times c}$. In the experiments, we fix the spatial resolution to be $R_x=R_y=128$, consistent with the majority of datasets considered. Data with different spatial resolutions are upsampled via interpolation or downsampled to match this resolution. To unify the number of channels (physical fields) across various PDEs, we zero-pad all input data to $c=4$ to match the data size with the largest number of channels.

Choosing a patch size of $P$ for each spatial dimension transforms each frame $\u(\cdot,t)$ into $(R_x/P) \times (R_y/P)$ patches, resulting in a sequence of up to $(T_0+T) \times (R_x/P) \times (R_y/P)$ visual tokens. We observed that the choice of patch size $P$ significantly influences the performance of the model, which is quantified in Section \ref{sec:ablation}. For our model, we select $P = 8$, the smallest size that maintains a reasonable sequence length. This choice results in a maximum sequence length of $(10+10) \times (128/8) \times (128/8) = 5120$.

After the transformer backbone, the sequence of tokens is projected back to physical space and then deflattened to reconstruct the predicted future states $\{\u(\cdot,t_i)\st T_0\le i<T_0+T\}$.

\subsection{Preliminary: Next Token Prediction}
In both language modeling and autoregressive image generation, next token prediction is the key learning task. For language models, the model predicts the probability distribution of the next token given the previous tokens, enabling the generation of coherent sentences by sampling tokens sequentially \cite{vaswani2017attention,radford2019language}. For autoregressive image generation, images are treated as sequences of pixels or patches, with each element predicted based on prior elements in the sequence \cite{chen2020generative}. Both applications rely on the autoregressive causal property, where future predictions depend only on prior context. More precisely, given a sequence of discrete tokens $X = (x_1,x_2,\dots,x_S)$, the modeling assumption for most language models is that: \begin{equation}\label{eqn:assum_lm}
    p(x_1,\dots,x_S) = \prod_{s=1}^S p(x_s|x_1,x_2,\dots,x_{s-1}).
\end{equation}
This is a reasonable assumption for sentence generation, as language typically follows a left-to-right sequential structure. However, for image generation, such dependencies are not as clear. Unlike text, images exhibit spatial invariance: the pixel/patch in the top-left corner does not inherently depend on the pixel/patch in the bottom-right corner.

\subsection{Next Frame Prediction}\label{sec:next_frame_prediction}
While the causal property is less intuitive for spatial dependencies, it is a natural assumption for temporal dynamics, especially for Markovian sequences. Next frame prediction is the natural solution that addresses these concerns, where visual tokens within the same frame can attend to each other and all tokens for the same frame will be generated in parallel conditioned on previous frames.

Denote the input token sequence as $X = (x_1,\dots,x_S)$ and the \model model as $f_\theta$, where $S = (T+T_0)\times N$ is the total sequence length and $N = (R_x/P) \times (R_y/P)$ is the number of tokens per frame. 
Instead of learning to predict the immediate next token, \[
    (x_1,x_2,\dots,x_s)\overset{f_\theta}{\longmapsto} (x_2, x_3,\dots, x_{s+1}),~~~\ ~~
\] the \model model learns to predict the next frame given the previous frames: \[
    (x_1,x_2,\dots,x_s)\overset{f_\theta}{\longmapsto} (x_{N+1}, x_{N+2}, \dots,x_{s+N}). 
\] 

To achieve this, similar to causal masks in transformers to mask out future tokens \cite{vaswani2017attention}, we use a block lower triangular causal mask \cite{cao2024vicon,tian2024visual} to allow spatial dependencies and temporal causality: \begin{equation}\label{eq:mask}
    M = \begin{bmatrix}
        \bm{1}_{N\times N} & \bm{0}_{N\times N} & \bm{0}_{N\times N} & \dots & \bm{0}_{N\times N} \\
        \bm{1}_{N\times N} & \bm{1}_{N\times N} & \bm{0}_{N\times N} & \dots & \bm{0}_{N\times N} \\
        \bm{1}_{N\times N} & \bm{1}_{N\times N} & \bm{1}_{N\times N} & \dots & \bm{0}_{N\times N} \\
        \vdots & \vdots & \vdots & \ddots & \vdots \\
        \bm{1}_{N\times N} & \bm{1}_{N\times N} & \bm{1}_{N\times N} & \dots & \bm{1}_{N\times N} \\
    \end{bmatrix}.
\end{equation}
Here $\bm{1}_{N\times N}$ and $\bm{0}_{N\times N}$ represent constant $1/0$ matrices of dimension $N\times N$. 

We compare next-frame prediction and next-token prediction in Section \ref{sec:ablation}, showing the accuracy gains from using next-frame prediction in the prediction tasks. For autoregressive inference, next-token prediction requires $T\times N$ prediction calls to generate $T$ frames, whereas next-frame prediction requires $T$ calls. Key-value caching further improves the inference efficiency; Section~\ref{sec:inference_speed} reports wall-clock measurements for this setting.

A similar block lower-triangular mask has been used for ``next function prediction'' and ``next scale prediction'' in VICON and VAR \cite{cao2024vicon,tian2024visual}. Our contribution is not the masking pattern itself: BCAT adapts this pattern to autoregressive two-dimensional fluid-dynamics forecasting, evaluates it against retrained PDE baselines, and studies its interaction with next-frame prediction and Muon optimization.

\paragraph{Difference from fixed-window update rules.} MPP \cite{mccabe2023multiple} and DPOT \cite{hao2024dpot} learn fixed-window update rules, whereas BCAT can condition on a flexible history window through temporal causality. This causal formulation permits parallel loss computation over future frames during training and supports key-value caching at inference.

\subsection{Muon Optimizer}
The Muon optimizer \cite{jordan2024muon} is a variant of SGD with preconditioned momentum. Muon is designed to optimize 2D parameters of the hidden layers, and the preconditioning step orthogonalizes the momentum matrix using Newton-Schulz matrix iteration \cite{schulz1933iterative}. When a network is trained with Muon optimizer, scalar/vector parameters of the network, as well as the input and output layers, should be optimized by a standard optimizer, such as AdamW \cite{loshchilovdecoupled}. 

Empirically, Muon demonstrated strong performance in training large language models \cite{jordan2024muon}, significantly outperforming AdamW in terms of convergence speed with $<1\%$ computational overhead. More recently, \cite{liu2025muon} optimizes Muon's distributed implementation and studies the scaling law for training larger LLMs with Muon optimizer. 
Observing the success of Muon optimizer in training LLMs, we train BCAT using a combination of Muon optimizer (for transformer 2D parameters) and AdamW optimizer, which significantly outperforms BCAT trained only with AdamW optimizer. BCAT is the first model using Muon optimizer for scientific computing and prediction tasks, and we observe that Muon optimizer is able to better approximate high frequency or high contrast information. The comparison is shown in Section \ref{sec:ablation}.

\subsection{Implementation Details}
The architecture utilizes a standard decoder-only transformer consisting of prenorm transformer blocks. We use RMS norm \cite{zhang2019root} in place of layer norms and add query-key normalizations \cite{henry2020query} after the attention projections, which we found to help stabilize the training process. In the feedforward network (FFN) of transformer layers, we use SwiGLU activation function \cite{shazeer2020glu}, and to keep a similar parameter count, the FFN hidden dimension is rescaled to be $\frac{8}{3}$x the feature dimension instead of the more common 4x. The patch embedding layer is efficiently implemented as a convolution layer with the same kernel size and stride (equivalent to a patchwise linear projection). The output reconstruction uses the corresponding convolutional projection, and inputs are zero-padded to four channels for a shared interface. More architecture and training details are included in Appendix~\ref{sec:exp_details}.

\section{Experiments} \label{sec:exp}

In this section, we first introduce the experiment setup and the baseline models we compare against. We present the main results comparing zero-shot performance in Section \ref{sec:main_results}, and then study a single in-distribution adaptation setting in Section \ref{sec:transfer}. To demonstrate the efficiency of BCAT, we compare inference speed and GPU memory usage in Section \ref{sec:inference_speed}. Various ablation studies for important architecture choices are included in Section \ref{sec:ablation}.
We include example visualizations of the BCAT model output and comparison of outputs from different models in Appendix~\ref{sec:more_visual}.

\subsection{Experiment Setup}
\paragraph{Dataset.}
The dataset we use consists of 6 parametric families of PDEs that model fluid dynamics in different regimes (some families contain subfamilies using different ranges of physical parameters, so different data distributions exist even within a family). We collect the dataset from 3 heterogeneous sources: PDEBench \cite{takamoto2022pdebench}, PDEArena \cite{gupta2022towards}, and CFDBench \cite{luo2023cfdbench}. The dataset includes shallow water equations and the Navier-Stokes system with incompressible and compressible flow, regular and complex geometries, and different buoyancy settings. For datasets that do not provide a train/val/test splitting, we use the standard 80\%/10\%/10\% splitting.  The total amount of data samples is about 69K trajectory sequences (time homogeneity can be used to create more training input/output pairs), which is the current amount used in comparable models. For a shared experimental interface, all data are resampled to $128\times128$ and retain their dataset-specific temporal resolutions. More details are included in Appendix~\ref{sec:dataset_details}.

\paragraph{Evaluation Metric.}
We use the relative $L^2$ norm as the evaluation metric. More precisely, given the ground truth $\u$ and model's prediction $\tilde{\u}$, we compute the (time-averaged) relative $L^2$ error: 
\begin{equation}
    \frac{1}{T}\sum_{i=T_0}^{T_0 + T-1}\frac{\|\u(\cdot,t_i) - \tilde{\u}(\cdot,t_i)\|_2}{\|\u(\cdot,t_i)\|_2 + \ep}  ,
\end{equation} 
where $T_0=10$ is the number of input steps, $T=10$ is the number of output steps, and $\ep = 10^{-7}$. 
Note that for the Navier-Stokes dataset from PDEArena, the temporal grid resolution is only 14, thus we set $T=4$ for this dataset only. The average used in the last column of Table~\ref{tab:main_results} is the average of the relative $L^2$ errors over the 6 families, i.e., the average over each row of the table.

\paragraph{Baselines and Comparisons.}
We evaluate our \model model against the following baseline methods. DeepONet \cite{lu2019deeponet} and FNO \cite{li2020fourier} are two widely used single-operator learning techniques designed to efficiently approximate solution operators for PDEs. UNet \cite{ronneberger2015u} is a classic convolutional neural network model for image processing, known for its symmetric hierarchical structure that enables it to capture both contextual information and fine details for pixel-wise predictions. 
ViT \cite{dosovitskiy2021an}, a transformer-based image processing model, is notable for its ability to capture global dependencies in images and its scalability across various image and video processing tasks. Lastly, we test against three recent PDE foundation models, PROSE-FD \cite{liu2024prosefd}, MPP \cite{mccabe2023multiple}, and DPOT \cite{hao2024dpot}. PROSE-FD fuses data input and symbolic equation encoding for multimodal operator learning. Both MPP and DPOT autoregressively predict PDE solutions, with MPP based on Axial-ViT and DPOT based on Adaptive Fourier Neural Operators. 

Additional details about these baselines can be found in Appendix \ref{sec:baseline_details}. We did not include Poseidon \cite{herde2024poseidon} as it does not generate meaningful results without fine-tuning, i.e., Poseidon cannot produce zero-shot predictions. Additionally, Poseidon cannot process history context, and thus is unable to handle partial observations of the state variables (e.g. the pressure and inhomogeneous forcing terms are not observed in PDEArena, but can be implicitly inferred from context). Some other transformer-based methods \cite{li2022transformer,wu2024transolver} are not considered as their model sizes are much smaller, both MPP and DPOT have comparable or larger model sizes than BCAT.
Note that DeepONet and FNO are designed for single tasks but can be extended to multi-task prediction using fine-tuning \cite{subramanian2024towards, zhang2024deeponet}. Except for the in-distribution adaptation in Section~\ref{sec:transfer}, all comparisons are performed in a zero-shot setting, which is also used by MPP and DPOT. As a reference, in Appendix \ref{sec:single_task_model}, we train separate single-task models for each family, and the results show no significant difference.

\subsection{Main Results}\label{sec:main_results}

We present the main results in Table \ref{tab:main_results}, where we report the relative $L^2$ error (\%) for each family of equations and their average. We use the same zero-shot setting for all models: a single model is trained to predict all families of equations without any fine-tuning. In this comparison, BCAT has the best average and the best result on five of six families. Appendix~\ref{sec:additional_exps} reports five-seed BCAT variability. On the more complicated PDEArena NS-cond dataset, BCAT has a large margin over the compared baselines.

\begin{figure*}[t]
    \centering
    \begin{subfigure}[b]{0.32\linewidth}
        \includegraphics[width=\linewidth]{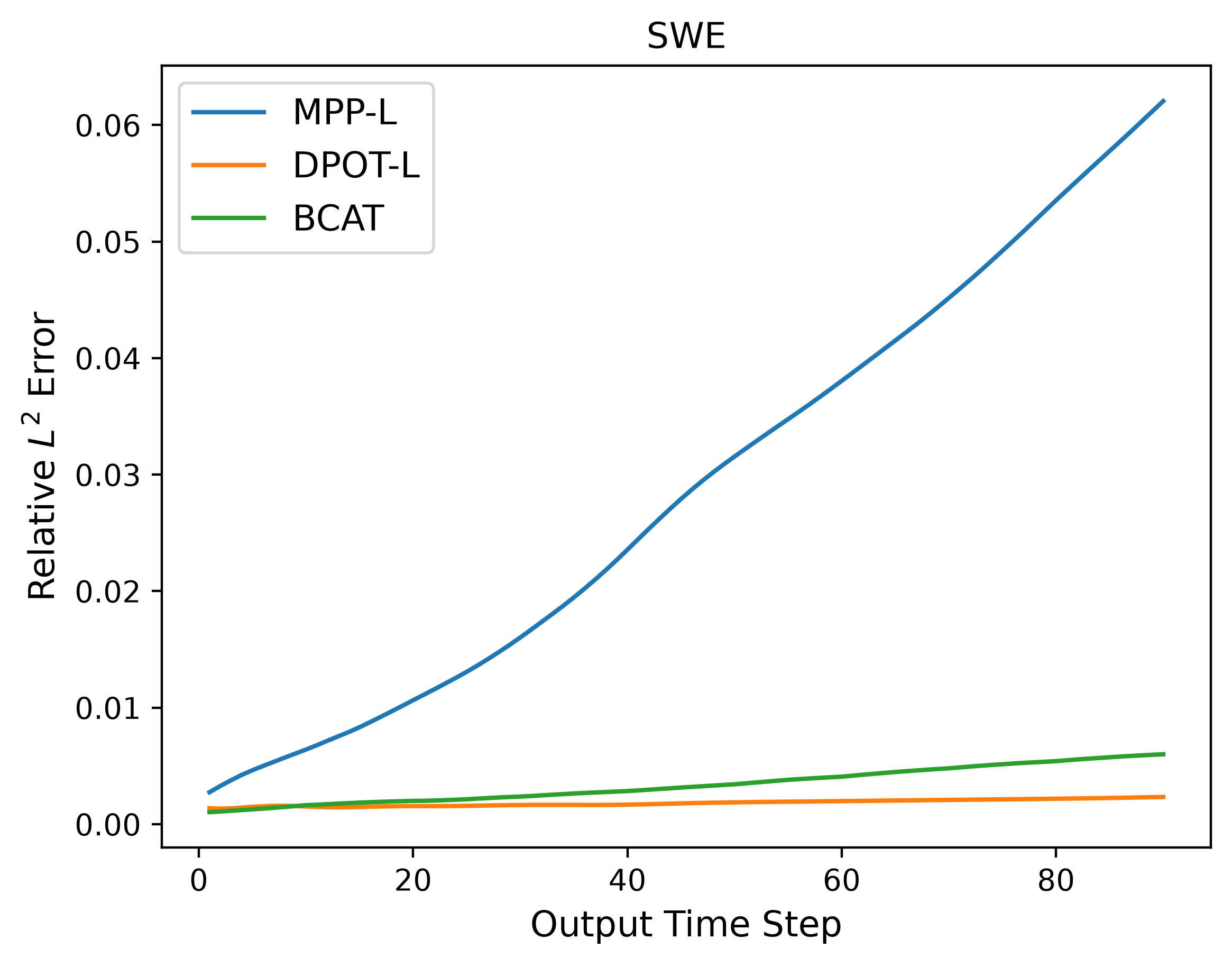}
        \caption{PDEBench SWE}
    \end{subfigure}\hfill
    \begin{subfigure}[b]{0.32\linewidth}
        \includegraphics[width=\linewidth]{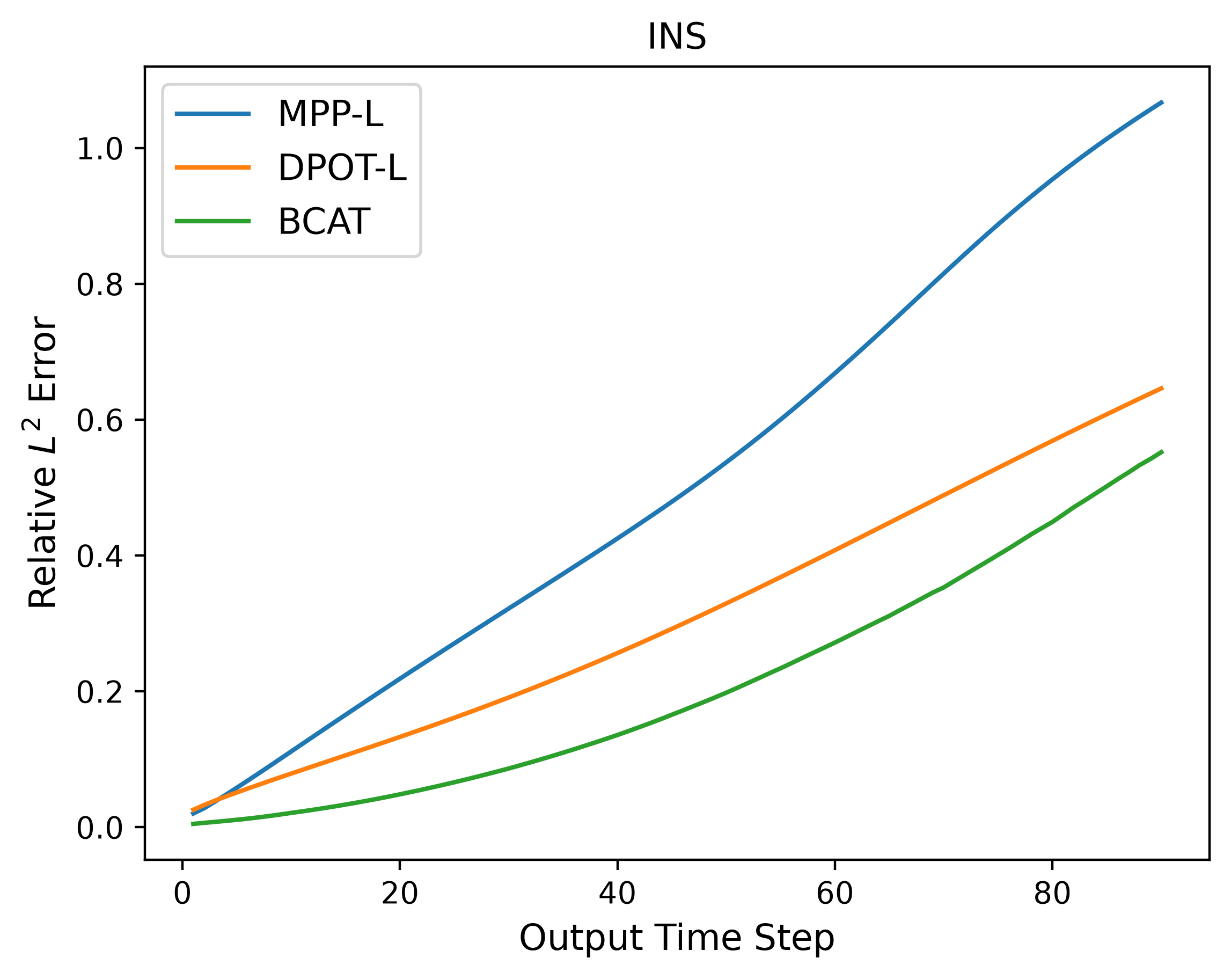}
        \caption{PDEBench INS}
    \end{subfigure}\hfill
    \begin{subfigure}[b]{0.32\linewidth}
        \includegraphics[width=\linewidth]{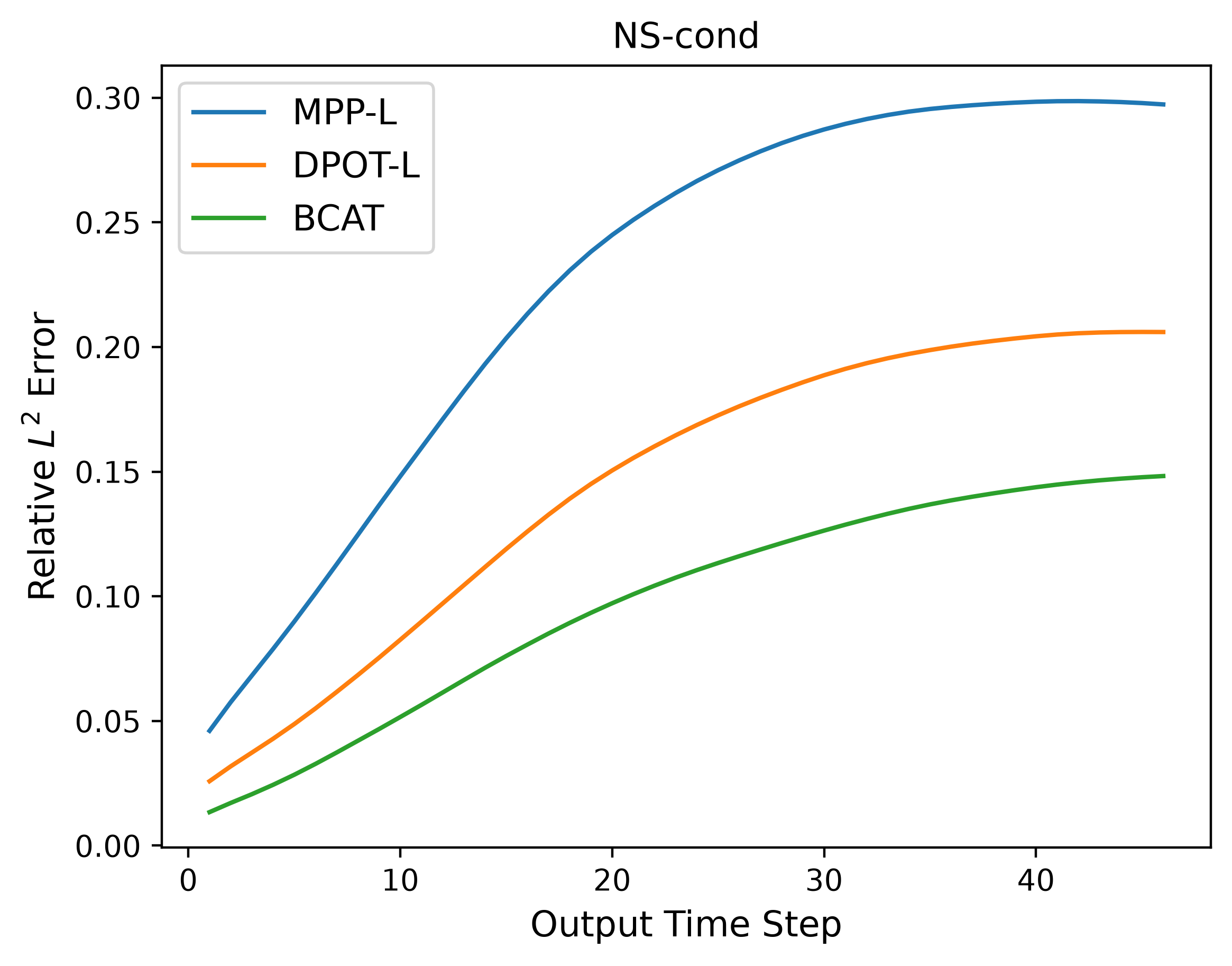}
        \caption{PDEArena NS-cond}
    \end{subfigure}
    \caption{\textbf{Long-horizon rollout error.} Relative $L^2$ error at each generated time step for the three datasets with trajectories long enough for extrapolative rollouts. All models use their generated outputs as subsequent inputs. The models are ordered from lower complexity (SWE) to higher complexity (NS). NS-cond plateaus late in the trajectory, whereas SWE and INS continue accumulating error.}
    \Description{Three line plots compare MPP-L, DPOT-L, and BCAT during long autoregressive rollouts. On shallow water, DPOT-L has the lowest error, BCAT remains low, and MPP-L increases sharply. On incompressible Navier-Stokes and conditioned Navier-Stokes, BCAT has the lowest error, followed by DPOT-L and then MPP-L.}
    \label{fig:long_rollouts}
\end{figure*}

We further study long-horizon autoregressive rollouts on longer datasets PDEBench SWE, PDEBench INS, and PDEArena NS-cond. For each, we recursively feed generated outputs back as inputs and compare BCAT with MPP-L and DPOT-L in Figure~\ref{fig:long_rollouts}. BCAT has the lowest per-step error for INS and NS-cond throughout the shown rollout; on SWE, DPOT-L remains slightly lower while both are substantially more stable than MPP-L. The trajectories have different dynamics: NS-cond approaches a plateau, whereas SWE and INS continue accumulating error.

\begin{table*}[t]
\centering
\caption{\textbf{Main Results and Comparisons with Baselines}. The numbers reported are relative $L^2$ errors (\%). The averages are taken with respect to the 6 distinct families listed in the columns of the table. We \textbf{bold} the best result in each column.
} 
\label{tab:main_results}
{
\begin{NiceTabular}{cc|cccccc|c}
\toprule
\multirow{2}{*}{Model} & \multirow{2}{*}{Param} & \multicolumn{3}{c|}{PDEBench}        & \multicolumn{2}{c|}{PDEArena}     & CFDBench & \multirow{2}{*}{Average} \\
                       &                        & SWE & CNS* & \multicolumn{1}{c|}{INS} & NS & \multicolumn{1}{c|}{NS-cond} & -      &                          \\ 
\midrule
DeepONet & 3.5M & 3.55 & 7.41 & 64.61 & 35.33 & 51.85 & 12.50 & 29.21 \\
FNO      & 0.6M & 3.71 & 6.31 & 36.84 & 38.67 & 55.63 & 8.52 & 24.95 \\
UNet     & 5.6M & 0.33 & 3.19 & 3.43 & 12.56  & 16.82 & 0.76 & 6.18 \\
MPP-B    & 116M & 1.02 & 1.90 & 7.52 & 5.71 & 12.57 & 1.23 & 4.99 \\
ViT & 162M & 0.25 & 1.49 & 2.82 & 7.05 & 12.41 & 0.55 & 4.10 \\
MPP-L    & 407M & 0.47 & 1.53 & 6.42 & 4.64 & 9.64 & 0.73 & 3.91 \\
DPOT-M     & 122M & 0.54 & {1.01} & 5.20 & {4.92} & {8.55} & 0.64 & {3.47} \\
PROSE-FD    & 165M & 0.28 & 1.41 & 2.75 & 5.27 & 9.61 & 0.61 & 3.32 \\ 
DPOT-L     & 523M & 0.15 & {0.89} & 4.08 & {2.21} & {5.29} & \textbf{0.34} & {2.16} \\
BCAT     & 156M & \textbf{0.10} & \textbf{0.39} & \textbf{1.34} & \textbf{1.59} & \textbf{3.13} & 0.52 & \textbf{1.18} \\

\bottomrule
\end{NiceTabular}
}
\end{table*}

\subsection{In-Distribution Adaptation}\label{sec:transfer}

We focus on and compare the zero-shot performance of the models considered, since zero-shot performance is an unbiased metric for the intrinsic capabilities of foundation models. For example, when GPT-2 \cite{radford2019language} was introduced, only the zero-shot performance was studied. Additionally, it may be difficult to directly compare PDE foundation models fine-tuned with different algorithms, such as full/partial parameter fine-tune and LoRA \cite{hu2022lora}. Fine-tuning could obscure a model's weakness through overfitting on task-specific objectives.

We evaluate low-data, in-distribution adaptation using the same full-parameter fine-tuning method for a fair comparison on the PDEBench Turbulence dataset, where the initial conditions lead to turbulent flow. We fine-tune pretrained BCAT, DPOT-L, and MPP-L for 500 gradient updates with a batch size of 64. Given the relatively small 1.6K-example training dataset, this is equivalent to 20 epochs and tests a low-data, low-compute regime. Table~\ref{tab:transfer} shows that BCAT outperforms the two larger baselines in this setting. Training BCAT from scratch for the same 500 steps gives 29.95\% relative $L^2$ error, demonstrating the benefit of pretraining.

BCAT's next-frame objective is well suited to this low-data adaptation setting. With a fixed compute budget, its causal formulation yields loss terms for multiple future frames in one forward pass, whereas the fixed-window MPP and DPOT updates yield one output step per pass. In this experiment, BCAT fine-tuning took about 10 minutes, compared with more than 20 minutes for MPP-L and DPOT-L.

\begin{table}[t]
    \centering
    \caption{\textbf{Comparing In-Distribution Adaptation on the PDEBench Turbulence Dataset.} Each model is fine-tuned for 500 gradient updates with a batch size of 64 (20 epochs).}
    \label{tab:transfer}
    \begin{NiceTabular}{cc|c}
        \toprule
        Model & Param & Testing Relative $L^2$ Error (\%) \\
        \midrule
        MPP-L & 407M &  7.35 \\
        DPOT-L & 523M & 6.76 \\
        BCAT & 156M & 3.87 \\
        \bottomrule
    \end{NiceTabular}
\end{table}

\subsection{Ablation Studies} \label{sec:ablation}
In this section, we present the results of our ablation studies to validate some key optimizer and architecture choices.

\paragraph{Muon Optimizer vs. AdamW Optimizer} We compare the BCAT model trained using a combination of Muon and AdamW optimizer (we refer to this as Muon for brevity) versus AdamW optimizer alone. When trained with Muon optimizer, the transformer layer matrix weights are optimized using Muon, and the other weights (scalar, vector, embedding, and initial/final projection layer weights) are optimized using AdamW, following \cite{jordan2024muon}. We kept the same optimizer hyperparameters (originally tuned for AdamW), and only tuned the learning rate scheduler hyperparameters. We found that more decay steps help stabilize the training process for Muon. 

\begin{figure*}[h!]
    \centering
    \includegraphics[width=0.95\linewidth]{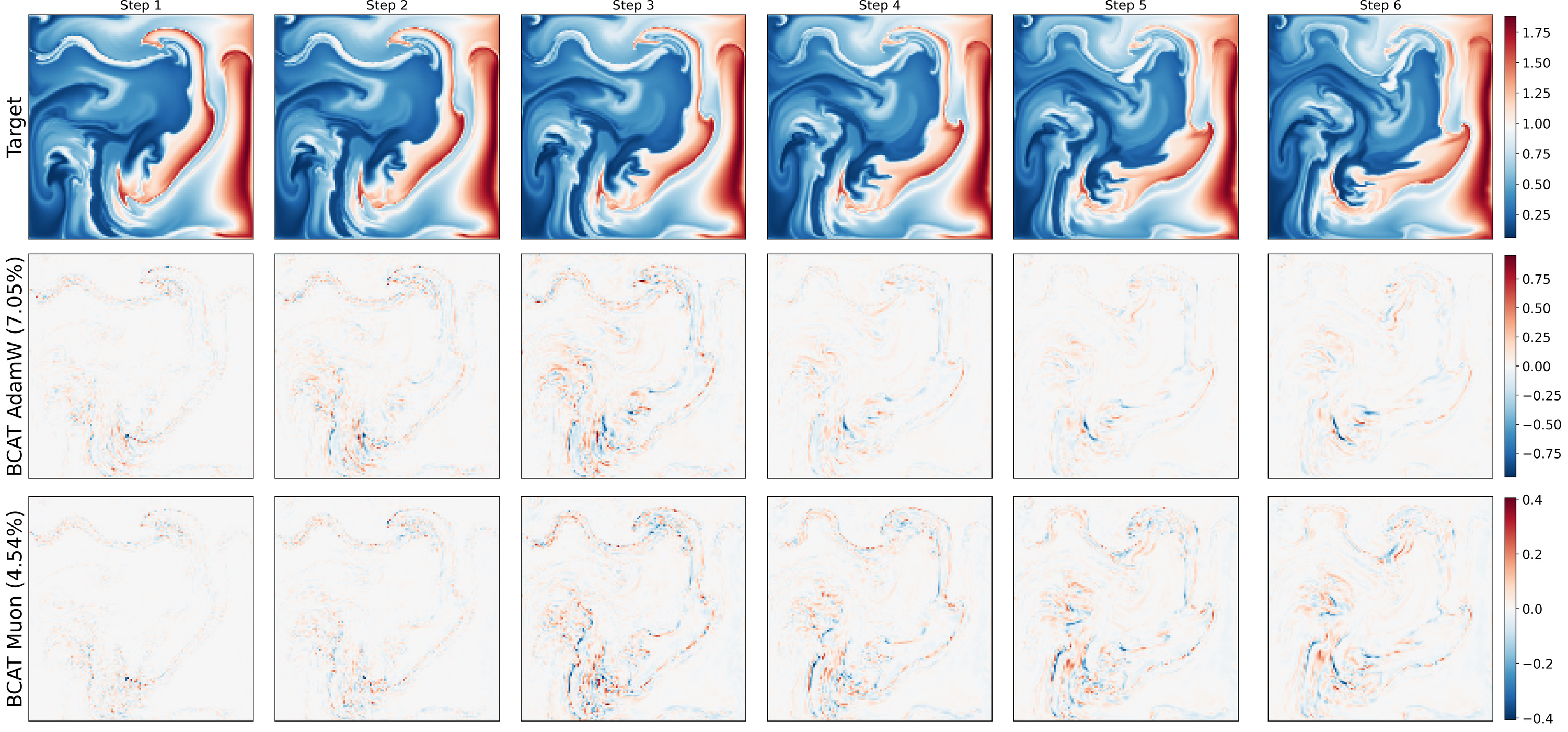}
\caption{\textbf{Comparing BCAT trained with Muon vs. AdamW optimizer.} The target (first row) is the first 6 output steps from the PDEArena Navier-Stokes (conditioned) dataset (particle density channel). For each optimizer (each row), we display the difference between the target and model output. Relative $L^2$ errors for the full trajectories are listed after the optimizer names.}
    \Description{A grid compares the ground-truth particle-density trajectory from conditioned Navier-Stokes with error maps from BCAT trained using AdamW and Muon. The Muon row has visibly lower error magnitude in high-contrast regions.}
    \label{fig:muon_adam}
\end{figure*}

The results are shown in Table~\ref{tab:optimizer}, where Muon outperforms AdamW in all datasets, improving BCAT's performance in all cases. In terms of average errors, Muon reduces the error from 1.92\% to 1.18\%, at the cost of <1\% computational overhead. We visualize and compare the output of BCAT trained using Muon and AdamW in Figure~\ref{fig:muon_adam}. We observe that, while the spatio-temporal error distributions are similar when trained with the two optimizers (in high contrast and high frequency regions), the error magnitude is much lower when trained with Muon. Appendix~\ref{sec:muon_diagnostics} further reports controlled convergence and frequency-domain diagnostics.

\begin{table*}[t!]
\centering
\caption{\textbf{Compare BCAT model trained with AdamW and Muon optimizer.} The numbers reported are relative $L^2$ errors (\%). The averages are taken with respect to the 6 distinct families listed in the columns of the table.
} 
\label{tab:optimizer}
{
\begin{NiceTabular}{cc|cccccc|c}
\toprule
\multirow{2}{*}{Optimizer} & \multirow{2}{*}{Param} & \multicolumn{3}{c|}{PDEBench}        & \multicolumn{2}{c|}{PDEArena}     & CFDBench & \multirow{2}{*}{Average} \\
                       &                        & SWE & CNS* & \multicolumn{1}{c|}{INS} & NS & \multicolumn{1}{c|}{NS-cond} & -      &                          \\ 
\midrule
AdamW     & 156M & {0.26} & {0.61} & {2.19} & {2.61} & {5.15} & 0.70 & {1.92} \\
Muon     & 156M & {0.10} & {0.39} & {1.34} & {1.59} & {3.13} & 0.52 & {1.18} \\
\bottomrule
\end{NiceTabular}
}
\end{table*}

\paragraph{Next Frame Prediction vs. Next Token Prediction.}
We directly compare the performance of \model with next frame prediction and next token prediction on the PDEArena NS dataset. We use the same model and training setup for both models, except that the next-token-prediction variant is trained to predict the next token instead of the next frame (and the mask is causal instead of block causal). When trained on a mixture of all 6 families, the next-token-prediction variant fails to converge, so we only train it on the PDEArena NS dataset (the shorter dataset with 14 frames).   

The results are shown in Table \ref{tab:frame_vs_token}. We observe that the next-frame-prediction model achieves a 3.5x better relative $L^2$ error compared to the next-token-prediction variant, demonstrating the effectiveness of our proposed training objective. We also compute the error on the training dataset and the ``next-token error.'' As the PDEArena NS dataset is a relatively small dataset, both models are overfitting to the training dataset to some degree; however, there is a performance difference between the two as shown by the error gap. 

The ``next-token error'' is computed assuming all previous tokens are exactly known (i.e. no prior error). More precisely, during testing time, we input the full trajectory to the model with proper masks to prevent data leaking, similar to training time. Both models are only asked to predict the immediate next token/frame and thus no autoregressive rollout is needed. We are thus measuring the amount of error over one step, i.e., the ``local error''. In this case, we observe that the performance of both models are relatively similar. This shows that the two models have similar (local) approximation power, but the next-frame-prediction model is better conditioned for multi-step prediction. Specifically, the next-frame-prediction has a slower accumulation of error rate than the next-token-prediction approach. As discussed in Section \ref{sec:next_frame_prediction}, to autoregressively generate $T$ future frames, the next-frame-prediction model only needs to be recursively called $T$ times, while the next-token-prediction variant needs to be recursively called $T\times N$ times (where $N$ is the number of tokens per frame). This may be the mechanism that leads to a significant reduction in error accumulation for the next-frame-prediction model.

\begin{table*}[t]
    \centering
    \caption{\textbf{Comparing next frame prediction and next token prediction.} The errors shown are relative $L^2$ errors (\%) for PDEArena NS dataset. *The Next-Token error is computed assuming all previous tokens are exactly known (i.e., the full trajectory is inputed to the model, similar to training time). It is not a standard metric and is only used to provide a comparison. }
    \label{tab:frame_vs_token}
    \begin{NiceTabular}{cc|c|c|c}
        \toprule
        Method & Param & Testing Error & Training Dataset Error & Next-Token Error* \\
        \midrule
        Next token prediction & 156M & 5.49 & 1.62 & 1.01 \\
        Next frame prediction & 156M & 1.59 & 0.91 & 0.81 \\
        \bottomrule
    \end{NiceTabular}
\end{table*}

\paragraph{Comparison with Spatio-Temporal Attention.}
The forward prediction task for 2D fluid dynamics is similar to a video generation task, where the state variable at each timestamp is akin to a frame in the video. In \cite{bertasius2021space}, several different forms of self-attention for processing spatial-temporal dynamics are compared for video classification task. Our \model model flattens the patches from all frames into a single sequence, and can be seen as a variant of joint space-time self-attention discussed in \cite{bertasius2021space}. Similarly, MPP \cite{mccabe2023multiple} is a variant of axial self-attention. Another approach that has not appeared in the literature as a backbone to PDE foundation models is the divided space-time self-attention, where temporal attention and spatial attention are separately applied one after the other. We implement this variant and refer to it as the Time-then-Space model. The temporal attention is a causal self-attention and the spatial attention is a full self-attention. 
All other configurations and hyperparameters are kept the same as the \model model. Due to more attention layers, the Time-then-Space model is larger than BCAT (206M vs 156M). The results are shown in Table~\ref{tab:st}, where the \model model slightly outperforms the spatio-temporal variant. Consider the difference in model sizes, this shows that the joint space-time self-attention is more effective for the forward prediction task.

\begin{table}[t]
    \centering
    \caption{\textbf{Comparing the performance of \model and Time-then-Space Model.} The errors shown are relative $L^2$ errors (\%).}
    \label{tab:st}
    \begin{NiceTabular}{cc|c}
        \toprule
        Model & Param & Average Testing Error \\
        \midrule
        Time-then-Space & 206M & 1.19 \\
        \model & 156M & 1.18 \\
        \bottomrule
    \end{NiceTabular}
\end{table}

\paragraph{Effect of Patch Size.}
In the image tokenization step, the patch size $P$ to divide the image into non-overlapping patches is a key hyperparameter. Note that this is a tradeoff between model complexity and feature spatial resolution, as half the patch size makes the sequence length 4x longer, thus 16x the time complexity and 4x space complexity (assuming memory efficient attention), even though the model parameter size remains similar, since models with smaller patch size have fewer parameters in patch embedding due to smaller dimensions per patch. However, we observe that having a smaller patch size and hence more visual tokens significantly improves the performance for both \model and ViT. The results are shown in Table \ref{tab:patch_size}. For patch size 4 or below, we refer to \cite{wang2025scaling} for a more comprehensive study on patch size scaling law (note that linear attention is used when the sequence length is beyond 4000).

\begin{table}[t]
    \centering
    \caption{\textbf{Comparing the performance of \model and ViT with different patch size.} The errors shown are relative $L^2$ errors (\%).}
    \label{tab:patch_size}
    \begin{NiceTabular}{ccc|c}
        \toprule
        Model & Patch Size & Param & Average Testing Error \\
        \midrule
         ViT & 16 & 162M & 4.10 \\
        ViT & 8 & 155M & 3.34 \\
        \model & 32 & 191M & 3.61 \\
        \model & 16 & 163M & 1.99 \\
        \model & 8 & 156M & 1.18 \\
        \bottomrule
    \end{NiceTabular}
\end{table}

\begin{table*}[h]
    \centering
    \caption{\textbf{Additional Ablation Studies.} Each row represents a different variant and we \underline{underline} the difference with the \model model (last row). The errors shown are relative $L^2$ errors (\%). *The training loss diverges when using Muon without QK-Norm, thus for this row only, we report the result when trained using AdamW.}
    \label{tab:ablation}
    \begin{NiceTabular}{cccc|c}
        \toprule
        Activation & QK-Norm & Mask & Param & Average Testing Error \\
        \midrule
        SwiGLU & \underline{\xmark} & block causal & 155.6M & 2.04* \\
        SwiGLU & \cmark  & \underline{causal}    & 155.6M & 1.97 \\
        \underline{GeLU} & \cmark & block causal & 154.8M & 1.34 \\
        SwiGLU & \cmark  & block causal & 155.6M & 1.18 \\
        \bottomrule
    \end{NiceTabular}
\end{table*}

\paragraph{Additional Architecture Ablation Study.}
We validate some architecture details by comparing the average testing error. The results are shown in Table \ref{tab:ablation}. We found query-key normalization \cite{henry2020query} helpful in stabilizing the training process, especially towards the end of the training phase. Notably, when the Muon optimizer is used, the training loss diverges without query-key normalization, demonstrating the effectiveness of this technique. This also suggests that the Muon optimizer has different optimization dynamics for query and key projection matrices. For this ablation experiment only, we report the result using the AdamW optimizer. Under similar parameter counts, we found SwiGLU activation function \cite{shazeer2020glu} better than GeLU. Note that the parameter count is not exactly the same since the MLP hidden dimension is rounded to the nearest multiple of 64. For the attention mask, to prevent data leaking in next frame prediction, any submask of the block causal mask \eqref{eq:mask} can work. We compare the \model model and a variant using a standard causal mask to perform next frame prediction. We observe that the block causal mask leads to better performance, demonstrating that having more token interactions within the same frame is beneficial for the forward prediction task. 

\subsection{Resource Usage Comparison} \label{sec:inference_speed}
In Table \ref{tab:inference}, we compare the inference speed and GPU memory usage for all models. To obtain these results, on a single NVIDIA H100 GPU, we use a batch size of 1 and measure the time required to generate 10 output steps for the shallow water equation. To ensure stability, we average run time over 50 repeats after some warmup. While smaller models (FNO, DeepONet, and UNet) have a faster run time, BCAT outperforms all models with comparable or larger sizes except ViT, which shares the same transformer layer without autoregressive rollout.

\begin{table}[t]
    \centering
    \caption{\textbf{Comparing the inference speed and GPU memory usage for all models.} We measure the resource usage for different models and variants to generate 10 output steps for a single trajectory.}
    \label{tab:inference}
    \begin{NiceTabular}{cc|cc}
        \toprule
        Model & Param & Inference Speed (ms) & GPU Memory Usage (MB) \\
        \midrule
        DeepONet & 3.5M & 2.7 & 190 \\
        FNO & 0.6M & 1.7 & 372 \\
        UNet & 5.6M & 1.3 & 135 \\
        \midrule
        ViT & 155M & 5.4 & 2362 \\
        MPP-B & 116M & 207.9 & 3087 \\
        MPP-L & 407M & 438.3 & 10869 \\
        DPOT-M & 122M & 95.7 & 1864 \\
        DPOT-L & 523M & 194.3 & 7985 \\
        \midrule
        Time-then-Space Variant & 206M & 175.5 & 3168 \\
        Next-token-prediction Variant & 156M & 13165 & 2402 \\
        BCAT (without KV cache) & 156M & 150.8 & 943 \\
        BCAT (with KV cache) & 156M & 70.9 & 2499 \\
        \bottomrule
    \end{NiceTabular}
\end{table}

\subsection{Experimental Summary and Model Novelty}
We summarize the experimental results and highlight the significance of the BCAT model. 

\paragraph{Performance.} BCAT improves the average zero-shot relative $L^2$ error in the six evaluated two-dimensional fluid families. Compared to the 3.4-times-larger DPOT-L baseline, BCAT reduces the average error by 45\%. It also performs well in low-data, in-distribution adaptation.

\paragraph{Learning objective.} BCAT's next-frame prediction adapts causal sequence modeling to spatiotemporal fluid data. It permits parallel training loss computation and, with a key-value cache, reduces the stated inference time from 150.8 to 70.9 ms for ten SWE output steps on one H100 GPU. The large speed gap with the next-token variant is enabled by predicting an entire frame at each autoregressive step.

\paragraph{Architecture improvements.} Architecture improvements in BCAT are helpful for the broader community. (1) In Section \ref{sec:ablation}, we show that decreasing the patch size improves the performance of both BCAT and ViT. (2) We drop-in replaced PROSE-FD's optimizer by Muon, and the performance improved from 3.32\% to 3.04\%. (3) Compared to standard activation functions like ReLU and GeLU, GLU variants have better smoothness and similar convergence speed. BCAT is the first to test and demonstrate their effectiveness in scientific computing. (4) BCAT shows the benefit of causality in terms of efficiency and performance. While models like MPP and DPOT are designed to be fixed window update rules, causality can also be integrated to improve their performance, similar to the Time-then-space variant of BCAT. 

\paragraph{Engineering considerations.} BCAT retains a standard decoder-only transformer backbone, which permits implementations using \texttt{bfloat16}, \texttt{torch.compile}, and \texttt{FlexAttention} \cite{dong2024flex}. On the stated hardware and configurations, BCAT trained in 42 hours, versus 120 hours for MPP-B and 80 hours for DPOT-M.

\section{Conclusion}

In this work, we introduced BCAT, a block-causal transformer for autoregressive prediction of two-dimensional fluid-dynamics solutions. Its next-frame objective permits spatial interactions within a frame while enforcing temporal causality across frames. Across six fluid families from PDEBench, PDEArena, and CFDBench, BCAT achieved the best average relative $L^2$ error in our zero-shot comparison while using fewer parameters than the largest baselines. The ablations show the benefit of next-frame prediction in the evaluated setting and of Muon relative to AdamW; the convergence and spectral analyses further characterize this optimizer effect.

The present evaluation focuses on two-dimensional fluid dynamics at a shared $128\times128$ resolution and dataset-specific temporal resolutions. Extending block-causal foundation models to additional PDE classes, higher-dimensional dynamics, new geometries, partial observations and variable temporal resolutions is a promising direction for future work.

\begin{acks}
Yuxuan Liu and Hayden Schaeffer are supported in part by NSF DMS 2427558. The work was supported by the NVIDIA Academic Grant Program 2026.
\end{acks}

\printbibliography

@article{liu2024prose,
  title={PROSE: Predicting Multiple Operators and Symbolic Expressions using multimodal transformers},
  author={Liu, Yuxuan and Zhang, Zecheng and Schaeffer, Hayden},
  journal={Neural Networks},
  volume={180},
  pages={106707},
  year={2024},
  publisher={Elsevier}
}

@article{zhang2024deeponet,
  title={DeepONet as a Multi-Operator Extrapolation Model: Distributed Pretraining with Physics-Informed Fine-Tuning},
  author={Zhang, Zecheng and Moya, Christian and Lu, Lu and Lin, Guang and Schaeffer, Hayden},
  journal={arXiv preprint arXiv:2411.07239},
  year={2024}
}

@article{jollie2024time,
  title={Time-series forecasting, knowledge distillation, and refinement within a multimodal pde foundation model},
  author={Jollie, Derek and Sun, Jingmin and Zhang, Zecheng and Schaeffer, Hayden},
  journal={arXiv preprint arXiv:2409.11609},
  year={2024}
}

@article{sun2024towards,
  title={Towards a Foundation Model for Partial Differential Equation: Multi-Operator Learning and Extrapolation},
  author={Sun, Jingmin and Liu, Yuxuan and Zhang, Zecheng and Schaeffer, Hayden},
  journal={arXiv preprint arXiv:2404.12355},
  year={2024}
}

@article{liu2024prosefd,
  title={PROSE-FD: A Multimodal PDE Foundation Model for Learning Multiple Operators for Forecasting Fluid Dynamics},
  author={Liu, Yuxuan and Sun, Jingmin and He, Xinjie and Pinney, Griffin and Zhang, Zecheng and Schaeffer, Hayden},
  journal={arXiv preprint arXiv:2409.09811},
  year={2024}
}

@article{sun2024lemon,
  title={LeMON: Learning to Learn Multi-Operator Networks},
  author={Sun, Jingmin and Zhang, Zecheng and Schaeffer, Hayden},
  journal={arXiv preprint arXiv:2408.16168},
  year={2024}
}

@article{pathak2022fourcastnet,
  title={Fourcastnet: A global data-driven high-resolution weather model using adaptive fourier neural operators},
  author={Pathak, Jaideep and Subramanian, Shashank and Harrington, Peter and Raja, Sanjeev and Chattopadhyay, Ashesh and Mardani, Morteza and Kurth, Thorsten and Hall, David and Li, Zongyi and Azizzadenesheli, Kamyar and others},
  journal={arXiv preprint arXiv:2202.11214},
  year={2022}
}

@article{herde2024poseidon,
  title={Poseidon: Efficient Foundation Models for PDEs},
  author={Herde, Maximilian and Raoni{\'c}, Bogdan and Rohner, Tobias and K{\"a}ppeli, Roger and Molinaro, Roberto and de B{\'e}zenac, Emmanuel and Mishra, Siddhartha},
  journal={arXiv preprint arXiv:2405.19101},
  year={2024}
}

@article{bodnar2024aurora,
  title={Aurora: A foundation model of the atmosphere},
  author={Bodnar, Cristian and Bruinsma, Wessel P and Lucic, Ana and Stanley, Megan and Brandstetter, Johannes and Garvan, Patrick and Riechert, Maik and Weyn, Jonathan and Dong, Haiyu and Vaughan, Anna and others},
  journal={arXiv preprint arXiv:2405.13063},
  year={2024}
}

@article{yang2024pde,
  title={PDE Generalization of In-Context Operator Networks: A Study on 1D Scalar Nonlinear Conservation Laws},
  author={Yang, Liu and Osher, Stanley J},
  journal={arXiv preprint arXiv:2401.07364},
  year={2024}
}

@article{yang2023prompting,
  title={Prompting in-context operator learning with sensor data, equations, and natural language},
  author={Yang, Liu and Meng, Tingwei and Liu, Siting and Osher, Stanley J},
  journal={arXiv preprint arXiv:2308.05061},
  year={2023}
}

@article{yang2023context,
  title={In-context operator learning with data prompts for differential equation problems},
  author={Yang, Liu and Liu, Siting and Meng, Tingwei and Osher, Stanley J},
  journal={Proceedings of the National Academy of Sciences},
  volume={120},
  number={39},
  pages={e2310142120},
  year={2023},
  publisher={National Acad Sciences}
}

@inproceedings{dosovitskiy2021an,
  title={An Image is Worth 16x16 Words: Transformers for Image Recognition at Scale},
  author={Alexey Dosovitskiy and Lucas Beyer and Alexander Kolesnikov and Dirk Weissenborn and Xiaohua Zhai and Thomas Unterthiner and Mostafa Dehghani and Matthias Minderer and Georg Heigold and Sylvain Gelly and Jakob Uszkoreit and Neil Houlsby},
  booktitle={International Conference on Learning Representations},
  year={2021},
}

@inproceedings{vaswani2017attention,
  author = {Vaswani, Ashish and Shazeer, Noam and Parmar, Niki and Uszkoreit, Jakob and Jones, Llion and Gomez, Aidan N. and Kaiser, \L{}ukasz and Polosukhin, Illia},
  title = {Attention is all you need},
  year = {2017},
  booktitle = {Proceedings of the 31st International Conference on Neural Information Processing Systems},
}

@inproceedings{chen2020generative,
  title={Generative pretraining from pixels},
  author={Chen, Mark and Radford, Alec and Child, Rewon and Wu, Jeffrey and Jun, Heewoo and Luan, David and Sutskever, Ilya},
  booktitle={International conference on machine learning},
  pages={1691--1703},
  year={2020},
  organization={PMLR}
}

@article{radford2019language,
  title={Language Models are Unsupervised Multitask Learners},
  author={Radford, Alec and Wu, Jeff and Child, Rewon and Luan, David and Amodei, Dario and Sutskever, Ilya},
  year={2019}
}

@article{cao2024vicon,
  title={VICON: Vision In-Context Operator Networks for Multi-Physics Fluid Dynamics Prediction},
  author={Cao, Yadi and Liu, Yuxuan and Yang, Liu and Yu, Rose and Schaeffer, Hayden and Osher, Stanley},
  journal={arXiv preprint arXiv:2411.16063},
  year={2024}
}

@article{tian2024visual,
  title={Visual autoregressive modeling: Scalable image generation via next-scale prediction},
  author={Tian, Keyu and Jiang, Yi and Yuan, Zehuan and Peng, Bingyue and Wang, Liwei},
  journal={arXiv preprint arXiv:2404.02905},
  year={2024}
}

@article{zhang2019root,
  title={Root mean square layer normalization},
  author={Zhang, Biao and Sennrich, Rico},
  journal={Advances in Neural Information Processing Systems},
  volume={32},
  year={2019}
}

@article{henry2020query,
  title={Query-key normalization for transformers},
  author={Henry, Alex and Dachapally, Prudhvi Raj and Pawar, Shubham and Chen, Yuxuan},
  journal={arXiv preprint arXiv:2010.04245},
  year={2020}
}

@article{takamoto2022pdebench,
  title={Pdebench: An extensive benchmark for scientific machine learning},
  author={Takamoto, Makoto and Praditia, Timothy and Leiteritz, Raphael and MacKinlay, Daniel and Alesiani, Francesco and Pfl{\"u}ger, Dirk and Niepert, Mathias},
  journal={Advances in Neural Information Processing Systems},
  volume={35},
  pages={1596--1611},
  year={2022}
}

@article{gupta2022towards,
  title={Towards Multi-spatiotemporal-scale Generalized PDE Modeling},
  author={Gupta, Jayesh K and Brandstetter, Johannes},
  journal={arXiv preprint arXiv:2209.15616},
  year={2022}
}

@article{luo2023cfdbench,
  title={Cfdbench: A comprehensive benchmark for machine learning methods in fluid dynamics},
  author={Luo, Yining and Chen, Yingfa and Zhang, Zhen},
  journal={arXiv preprint arXiv:2310.05963},
  year={2023}
}

@article{dubey2024llama,
  title={The llama 3 herd of models},
  author={Dubey, Abhimanyu and Jauhri, Abhinav and Pandey, Abhinav and Kadian, Abhishek and Al-Dahle, Ahmad and Letman, Aiesha and Mathur, Akhil and Schelten, Alan and Yang, Amy and Fan, Angela and others},
  journal={arXiv preprint arXiv:2407.21783},
  year={2024}
}

@article{shazeer2020glu,
  title={Glu variants improve transformer},
  author={Shazeer, Noam},
  journal={arXiv preprint arXiv:2002.05202},
  year={2020}
}

@article{li2020fourier,
  title={Fourier neural operator for parametric partial differential equations},
  author={Li, Zongyi and Kovachki, Nikola and Azizzadenesheli, Kamyar and Liu, Burigede and Bhattacharya, Kaushik and Stuart, Andrew and Anandkumar, Anima},
  journal={arXiv preprint arXiv:2010.08895},
  year={2020}
}

@article{lu2019deeponet,
  title={Deeponet: Learning nonlinear operators for identifying differential equations based on the universal approximation theorem of operators},
  author={Lu, Lu and Jin, Pengzhan and Karniadakis, George Em},
  journal={arXiv preprint arXiv:1910.03193},
  year={2019}
}

@inproceedings{ronneberger2015u,
  title={U-net: Convolutional networks for biomedical image segmentation},
  author={Ronneberger, Olaf and Fischer, Philipp and Brox, Thomas},
  booktitle={Medical image computing and computer-assisted intervention--MICCAI 2015: 18th international conference, Munich, Germany, October 5-9, 2015, proceedings, part III 18},
  pages={234--241},
  year={2015},
  organization={Springer}
}

@article{mccabe2023multiple,
  title={Multiple physics pretraining for physical surrogate models},
  author={McCabe, Michael and Blancard, Bruno R{\'e}galdo-Saint and Parker, Liam Holden and Ohana, Ruben and Cranmer, Miles and Bietti, Alberto and Eickenberg, Michael and Golkar, Siavash and Krawezik, Geraud and Lanusse, Francois and others},
  journal={arXiv preprint arXiv:2310.02994},
  year={2023}
}

@article{hao2024dpot,
  title={Dpot: Auto-regressive denoising operator transformer for large-scale pde pre-training},
  author={Hao, Zhongkai and Su, Chang and Liu, Songming and Berner, Julius and Ying, Chengyang and Su, Hang and Anandkumar, Anima and Song, Jian and Zhu, Jun},
  journal={arXiv preprint arXiv:2403.03542},
  year={2024}
}

@inproceedings{bertasius2021space,
  title={Is space-time attention all you need for video understanding?},
  author={Bertasius, Gedas and Wang, Heng and Torresani, Lorenzo},
  booktitle={ICML},
  volume={2},
  number={3},
  pages={4},
  year={2021}
}

@article{hu2024minicpm,
  title={Minicpm: Unveiling the potential of small language models with scalable training strategies},
  author={Hu, Shengding and Tu, Yuge and Han, Xu and He, Chaoqun and Cui, Ganqu and Long, Xiang and Zheng, Zhi and Fang, Yewei and Huang, Yuxiang and Zhao, Weilin and others},
  journal={arXiv preprint arXiv:2404.06395},
  year={2024}
}

@article{radford2018improving,
  title={Improving language understanding by generative pre-training},
  author={Radford, Alec and Narasimhan, Karthik and Salimans, Tim and Sutskever, Ilya and others},
  year={2018},
  publisher={OpenAI}
}

@article{brown2020language,
  title={Language models are few-shot learners},
  author={Brown, Tom and Mann, Benjamin and Ryder, Nick and Subbiah, Melanie and Kaplan, Jared D and Dhariwal, Prafulla and Neelakantan, Arvind and Shyam, Pranav and Sastry, Girish and Askell, Amanda and others},
  journal={Advances in neural information processing systems},
  volume={33},
  pages={1877--1901},
  year={2020}
}

@inproceedings{ramesh2021zero,
  title={Zero-shot text-to-image generation},
  author={Ramesh, Aditya and Pavlov, Mikhail and Goh, Gabriel and Gray, Scott and Voss, Chelsea and Radford, Alec and Chen, Mark and Sutskever, Ilya},
  booktitle={International conference on machine learning},
  pages={8821--8831},
  year={2021},
  organization={Pmlr}
}

@article{ramesh2022hierarchical,
  title={Hierarchical text-conditional image generation with clip latents},
  author={Ramesh, Aditya and Dhariwal, Prafulla and Nichol, Alex and Chu, Casey and Chen, Mark},
  journal={arXiv preprint arXiv:2204.06125},
  volume={1},
  number={2},
  pages={3},
  year={2022}
}

@inproceedings{rombach2022high,
  title={High-resolution image synthesis with latent diffusion models},
  author={Rombach, Robin and Blattmann, Andreas and Lorenz, Dominik and Esser, Patrick and Ommer, Bj{\"o}rn},
  booktitle={Proceedings of the IEEE/CVF conference on computer vision and pattern recognition},
  pages={10684--10695},
  year={2022}
}

@article{touvron2023llama,
  title={Llama: Open and efficient foundation language models},
  author={Touvron, Hugo and Lavril, Thibaut and Izacard, Gautier and Martinet, Xavier and Lachaux, Marie-Anne and Lacroix, Timoth{\'e}e and Rozi{\`e}re, Baptiste and Goyal, Naman and Hambro, Eric and Azhar, Faisal and others},
  journal={arXiv preprint arXiv:2302.13971},
  year={2023}
}

@article{touvron2023llama2,
  title={Llama 2: Open foundation and fine-tuned chat models},
  author={Touvron, Hugo and Martin, Louis and Stone, Kevin and Albert, Peter and Almahairi, Amjad and Babaei, Yasmine and Bashlykov, Nikolay and Batra, Soumya and Bhargava, Prajjwal and Bhosale, Shruti and others},
  journal={arXiv preprint arXiv:2307.09288},
  year={2023}
}

@article{bommasani2021opportunities,
  title={On the opportunities and risks of foundation models},
  author={Bommasani, Rishi and Hudson, Drew A and Adeli, Ehsan and Altman, Russ and Arora, Simran and von Arx, Sydney and Bernstein, Michael S and Bohg, Jeannette and Bosselut, Antoine and Brunskill, Emma and others},
  journal={arXiv preprint arXiv:2108.07258},
  year={2021}
}

@article{achiam2023gpt,
  title={Gpt-4 technical report},
  author={Achiam, Josh and Adler, Steven and Agarwal, Sandhini and Ahmad, Lama and Akkaya, Ilge and Aleman, Florencia Leoni and Almeida, Diogo and Altenschmidt, Janko and Altman, Sam and Anadkat, Shyamal and others},
  journal={arXiv preprint arXiv:2303.08774},
  year={2023}
}

@article{yang2024qwen2,
  title={Qwen2. 5 Technical Report},
  author={Yang, An and Yang, Baosong and Zhang, Beichen and Hui, Binyuan and Zheng, Bo and Yu, Bowen and Li, Chengyuan and Liu, Dayiheng and Huang, Fei and Wei, Haoran and others},
  journal={arXiv preprint arXiv:2412.15115},
  year={2024}
}

@misc{brooks2024video,
  title={Video generation models as world simulators},
  author={Brooks, Tim and Peebles, Bill and Holmes, Connor and DePue, Will and Guo, Yufei and Jing, Li and Schnurr, David and Taylor, Joe and Luhman, Troy and Luhman, Eric and others},
  journal={2024-03-03]. https://openai. com/research/video-generation-modelsas-world-simulators},
  year={2024}
}

@article{subramanian2024towards,
  title={Towards foundation models for scientific machine learning: Characterizing scaling and transfer behavior},
  author={Subramanian, Shashank and Harrington, Peter and Keutzer, Kurt and Bhimji, Wahid and Morozov, Dmitriy and Mahoney, Michael W and Gholami, Amir},
  journal={Advances in Neural Information Processing Systems},
  volume={36},
  year={2024}
}

@article{jin2023time,
  title={Time-llm: Time series forecasting by reprogramming large language models},
  author={Jin, Ming and Wang, Shiyu and Ma, Lintao and Chu, Zhixuan and Zhang, James Y and Shi, Xiaoming and Chen, Pin-Yu and Liang, Yuxuan and Li, Yuan-Fang and Pan, Shirui and others},
  journal={arXiv preprint arXiv:2310.01728},
  year={2023}
}

@article{zhou2024can,
  title={Can LLMs Understand Time Series Anomalies?},
  author={Zhou, Zihao and Yu, Rose},
  journal={arXiv preprint arXiv:2410.05440},
  year={2024}
}

@article{liu2021gated,
  title={Gated transformer networks for multivariate time series classification},
  author={Liu, Minghao and Ren, Shengqi and Ma, Siyuan and Jiao, Jiahui and Chen, Yizhou and Wang, Zhiguang and Song, Wei},
  journal={arXiv preprint arXiv:2103.14438},
  year={2021}
}

@article{gruver2024large,
  title={Large language models are zero-shot time series forecasters},
  author={Gruver, Nate and Finzi, Marc and Qiu, Shikai and Wilson, Andrew G},
  journal={Advances in Neural Information Processing Systems},
  volume={36},
  year={2024}
}

@article{chang2024llm4ts,
  title={Llm4ts: Aligning pre-trained llms as data-efficient time-series forecasters},
  author={Chang, Ching and Wang, Wei-Yao and Peng, Wen-Chih and Chen, Tien-Fu},
  journal={arXiv preprint arXiv:2308.08469},
  year={2024}
}

@inproceedings{tan2024language,
  title={Are language models actually useful for time series forecasting?},
  author={Tan, Mingtian and Merrill, Mike A and Gupta, Vinayak and Althoff, Tim and Hartvigsen, Thomas},
  booktitle={The Thirty-eighth Annual Conference on Neural Information Processing Systems},
  year={2024}
}

@article{liu2024can,
  title={How can large language models understand spatial-temporal data?},
  author={Liu, Lei and Yu, Shuo and Wang, Runze and Ma, Zhenxun and Shen, Yanming},
  journal={arXiv preprint arXiv:2401.14192},
  year={2024}
}

@article{abdin2024phi3,
  title={Phi-3 technical report: A highly capable language model locally on your phone},
  author={Abdin, Marah and Aneja, Jyoti and Awadalla, Hany and Awadallah, Ahmed and Awan, Ammar Ahmad and Bach, Nguyen and Bahree, Amit and Bakhtiari, Arash and Bao, Jianmin and Behl, Harkirat and others},
  journal={arXiv preprint arXiv:2404.14219},
  year={2024}
}

@article{abdin2024phi4,
  title={Phi-4 technical report},
  author={Abdin, Marah and Aneja, Jyoti and Behl, Harkirat and Bubeck, S{\'e}bastien and Eldan, Ronen and Gunasekar, Suriya and Harrison, Michael and Hewett, Russell J and Javaheripi, Mojan and Kauffmann, Piero and others},
  journal={arXiv preprint arXiv:2412.08905},
  year={2024}
}

@article{team2024gemma,
  title={Gemma 2: Improving open language models at a practical size},
  author={Team, Gemma and Riviere, Morgane and Pathak, Shreya and Sessa, Pier Giuseppe and Hardin, Cassidy and Bhupatiraju, Surya and Hussenot, L{\'e}onard and Mesnard, Thomas and Shahriari, Bobak and Ram{\'e}, Alexandre and others},
  journal={arXiv preprint arXiv:2408.00118},
  year={2024}
}

@misc{jordan2024muon,
  author       = {Keller Jordan and Yuchen Jin and Vlado Boza and Jiacheng You and
                  Franz Cesista and Laker Newhouse and Jeremy Bernstein},
  title        = {Muon: An optimizer for hidden layers in neural networks},
  year         = {2024},
  url          = {https://kellerjordan.github.io/posts/muon/}
}

@article{liu2025muon,
  title={Muon is scalable for llm training},
  author={Liu, Jingyuan and Su, Jianlin and Yao, Xingcheng and Jiang, Zhejun and Lai, Guokun and Du, Yulun and Qin, Yidao and Xu, Weixin and Lu, Enzhe and Yan, Junjie and others},
  journal={arXiv preprint arXiv:2502.16982},
  year={2025}
}

@inproceedings{loshchilovdecoupled,
  title={Decoupled Weight Decay Regularization},
  author={Loshchilov, Ilya and Hutter, Frank},
  booktitle={International Conference on Learning Representations}
}

@article{hu2022lora,
  title={Lora: Low-rank adaptation of large language models.},
  author={Hu, Edward J and Shen, Yelong and Wallis, Phillip and Allen-Zhu, Zeyuan and Li, Yuanzhi and Wang, Shean and Wang, Lu and Chen, Weizhu and others},
  journal={ICLR},
  volume={1},
  number={2},
  pages={3},
  year={2022}
}

@article{schulz1933iterative,
  title={Iterative berechung der reziproken matrix},
  author={Schulz, G{\"u}nther},
  journal={ZAMM-Journal of Applied Mathematics and Mechanics/Zeitschrift f{\"u}r Angewandte Mathematik und Mechanik},
  volume={13},
  number={1},
  pages={57--59},
  year={1933},
  publisher={Wiley Online Library}
}

@article{wang2025scaling,
  title={Scaling Laws in Patchification: An Image Is Worth 50,176 Tokens And More},
  author={Wang, Feng and Yu, Yaodong and Wei, Guoyizhe and Shao, Wei and Zhou, Yuyin and Yuille, Alan and Xie, Cihang},
  journal={arXiv preprint arXiv:2502.03738},
  year={2025}
}

@article{wu2024transolver,
  title={Transolver: A fast transformer solver for pdes on general geometries},
  author={Wu, Haixu and Luo, Huakun and Wang, Haowen and Wang, Jianmin and Long, Mingsheng},
  journal={arXiv preprint arXiv:2402.02366},
  year={2024}
}

@article{li2022transformer,
  title={Transformer for partial differential equations' operator learning},
  author={Li, Zijie and Meidani, Kazem and Farimani, Amir Barati},
  journal={arXiv preprint arXiv:2205.13671},
  year={2022}
}

@article{dong2024flex,
  title={Flex Attention: A Programming Model for Generating Optimized Attention Kernels},
  author={Dong, Juechu and Feng, Boyuan and Guessous, Driss and Liang, Yanbo and He, Horace},
  journal={arXiv preprint arXiv:2412.05496},
  year={2024}
}

\appendix

\section{Dataset Details} \label{sec:dataset_details}
The data was obtained from the PDEBench \cite{takamoto2022pdebench}, PDEArena \cite{gupta2022towards}, and CFDBench \cite{luo2023cfdbench} datasets. Unless otherwise specified, the space resolution is $128\times 128$. 

\subsection{PDEBench}
\paragraph{Shallow Water Equation.}
The quantity of interest is the water depth $h(\x,t)$ on domain $[-2.5,2.5]^2\times [0,1]$ with Neumann boundary condition. The temporal resolution is 101. The equations are: \begin{align}
    \p_t h + \nabla h\u &= 0,\\
    \p_t h\u + \nabla \left(h \,\u \cdot \u + \frac12 g_r h^2\right) &= - g_r h \nabla b.
\end{align}

\paragraph{Incompressible Navier-Stokes Equation.}
The quantities of interest are the velocities $\u(\x,t)$ and particle density $c(\x,t)$ on domain $[0,1]^2\times [0,5]$ with Dirichlet boundary condition. The temporal resolution is 1000. The equations are: 
\begin{align}
    \rho(\p_t \u + \u \cdot \nabla \u) &= -\nabla p + \mu \Delta \u + \mathbf{F},\\
    \nabla \cdot \u &= 0,\\
    \p_t c + \nabla \cdot (c\u) &= 0.
\end{align} 
The forcing term $\mathbf{F}$ is randomly sampled.  

\paragraph{Compressible Navier-Stokes Equation.}
The quantities of interest are the velocities $\u(\x,t)$, pressure $p(\x,t)$, and density $\rho(\x,t)$ on domain $[0,1]^2\times [0,1]$ with periodic boundary conditions. The temporal resolution is 21. For equations with low viscosities, the dataset is provided on a finer $512\times 512$ space grid, which is downsampled to $128\times 128$ for consistency (through average pooling). The equations are:
\begin{align}\label{eq:cns}
    \partial_t \rho + \nabla \cdot (\rho \u) &= 0,\\
    \rho(\partial_t \u + \u\cdot \nabla \u) &= - \nabla p + \eta \Delta \u + (\zeta + \eta/3) \nabla (\nabla\cdot \u),\\
    \partial_t \left(\ep + \frac{\rho u^2}{2}\right) &= - \nabla \cdot \left(\left(\varepsilon + p + \frac{\rho u^2}{2}\right)\u - \u\cdot \sigma'\right).
\end{align}

\subsection{PDEArena}
\paragraph{Incompressible Navier-Stokes Equation.}
The quantities of interest are the velocities $\u(\x,t)$ and particle density $c(\x,t)$ on domain $[0,32]^2\times [18,102]$ with Dirichlet boundary conditions for velocity and Neumann boundary condition for particle field. The temporal resolution is 14. The equations are: \begin{align}\label{eq:arena_ns}
    \rho(\p_t \u + \u \cdot \nabla \u) &= -\nabla p + \mu \Delta \u + \mathbf{F},\\
    \nabla \cdot \u &= 0,\\
    \p_t c + \nabla \cdot (c\u) &= 0.
\end{align} The forcing term $\mathbf{F}$ takes the form $\mathbf{F} = (0, f)$ with $f=0.5$. 

\paragraph{Incompressible Navier-Stokes Equation (Conditioned).}
The quantities of interest are the velocities $\u(\x,t)$ and particle density $c(\x,t)$ on domain $[0,32]^2\times [18,102]$ with Dirichlet boundary conditions for velocity and Neumann boundary condition for particle field. The temporal resolution is 56. The equations are: \begin{align}\label{eq:arena_ns_c}
    \rho(\p_t \u + \u \cdot \nabla \u) &= -\nabla p + \mu \Delta \u + \mathbf{F},\\
    \nabla \cdot \u &= 0,\\
    \p_t c + \nabla \cdot (c\u) &= 0.
\end{align} The forcing term $\mathbf{F}$ takes the form $\mathbf{F} = (0, f)$ where $f$ is uniformly sampled in $[0.2,0.5]$. 

\subsection{CFDBench}
\paragraph{Incompressible Navier-Stokes Equation.}
The quantities of interest are the velocities $\u(\x,t)$ and pressure $p(\x,t)$. This dataset contains irregular geometries with Dirichlet boundary conditions. The raw space resolution is $64\times 64$ which is upsampled to $128\times 128$ via interpolation. The equations are: \begin{align}
    \rho(\p_t \u + \u \cdot \nabla \u) &= -\nabla p + \mu \Delta \u,\\
    \nabla \cdot \u &= 0.
\end{align}

\section{Experiment Details} \label{sec:exp_details}

\subsection{Training}\label{sec:train_details}
We perform data normalization during the training process. Given the input sequence of data $\{\u(\cdot,t_i)\st 0\le i < T_0\}$, we compute the mean and standard deviation of each input trajectory, which are used to normalize both the input and ground truth sequence. The loss function is the standard mean squared error in the normalized space. The models are trained using a combination of Muon optimizer \cite{jordan2024muon,liu2025muon} and AdamW optimizer \cite{loshchilovdecoupled} with a global batch size of 128 for 40 epochs where each epoch is 4,000 steps. The Muon optimizer is applied to all transformer matrix parameters, while AdamW optimizer is applied to the rest of the parameters (scalar, embedding, initial and final projection layer). The warmup-stable-decay learning rate scheduler \cite{hu2024minicpm} is used with 10\% warmup and 50\% decay. We use learning rate $10^{-3}$ and weight decay $10^{-4}$. On four NVIDIA H100 GPUs, the training takes about 42 hours.

\subsection{Model Hyperparameters}\label{sec:model_details}
\begin{figure}[h]
    \centering
    \includegraphics[width=0.25\linewidth]{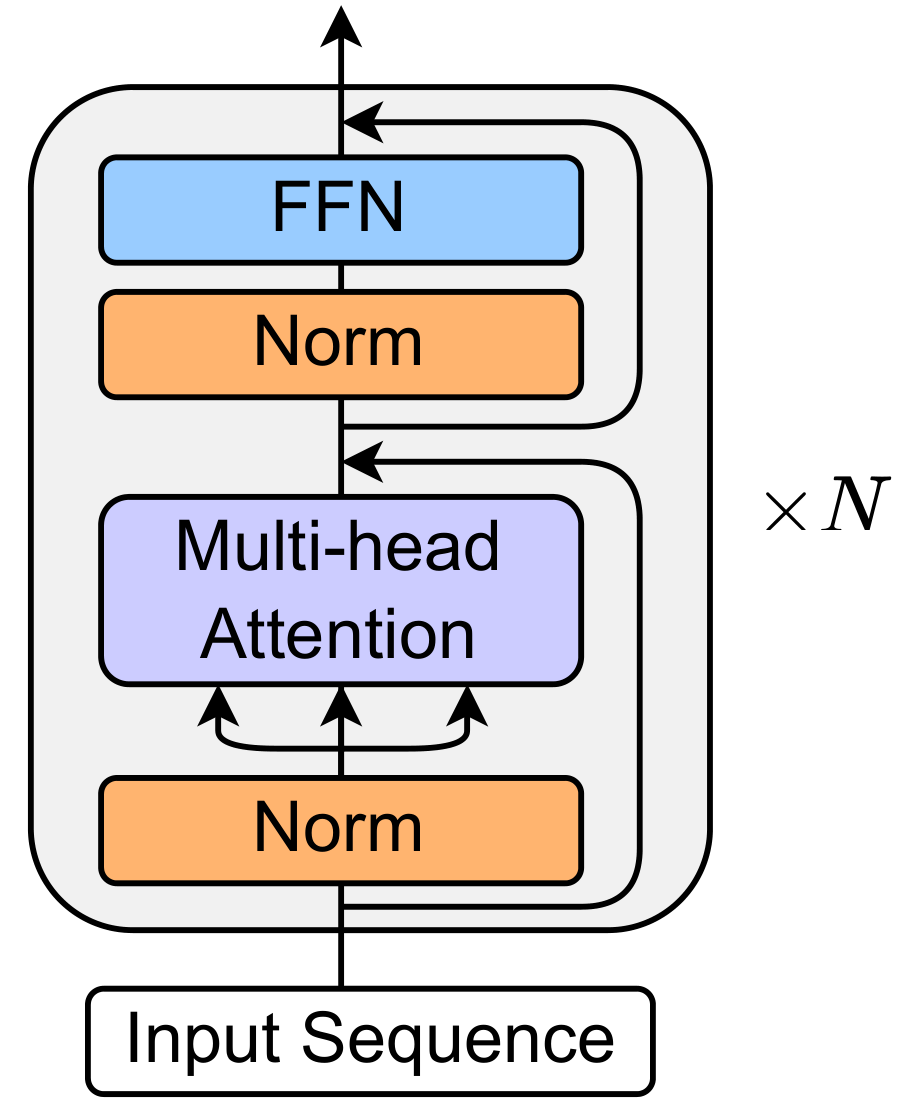}
\caption{Transformer Layers used in \model model.}
    \Description{A schematic of one prenorm decoder-only transformer layer used by BCAT, containing RMS normalization, multi-head self-attention with query-key normalization, residual connections, and a SwiGLU feedforward network.}
    \label{fig:layer}
\end{figure}

We summarize the model hyperparameters in Table~\ref{tab:model_hyper}. An illustration of a transformer layer used in \model model is shown in Figure~\ref{fig:layer}.

\begin{table}[h]
    \centering
    \footnotesize
    \caption{\textbf{Model hyperparameters.} FFN means feedforward network.}
    \label{tab:model_hyper}
    \begin{tabular}{l l | l l }
    \toprule
    Hidden dimension - attention & 1024 & Hidden dimension - FFN & 2752 \\
    Number of attention heads & 8 & Number of layers & 12\\
    Activation & SwiGLU & Dropout & 0 \\
    Patch Size & 8 &  &  \\
    \bottomrule
    \end{tabular}
\end{table}

\subsection{Baselines and Comparisons}\label{sec:baseline_details}
In this section, we include more details about the compared models.

\paragraph{DeepONet \cite{lu2019deeponet}.}
We employ the unstacked DeepONet architecture, consisting of a single trunk network and a single branch network. Initially, the input data is divided into $8 \times 8$ patches, with each patch being embedded into a 128-dimensional vector. These vectors are then passed through the branch network, producing an output with a basis dimension of $p = 50$. Simultaneously, the query point is processed through the trunk network, which also outputs a vector with the same dimension, $p$. The output solution at the query point is obtained by taking the inner product of the outputs from the two networks.

\paragraph{FNO \cite{li2020fourier}.}
We use 4 layers of standard 3d FNO to process the input data. The number of modes to keep in each dimension is set to 8, and the number of hidden channels is set to 16. The 3d FNO model directly maps 10 input steps to 10 output steps.

\paragraph{UNet \cite{ronneberger2015u}.}
We use 8 layers of 3d UNet with GeLU activation and 32 hidden dimensions. The 3d UNet model directly maps 10 input steps to 10 output steps (Padding is added to the input to make it a power of 2, which is needed for the shrinkage in dimension).

\paragraph{ViT \cite{dosovitskiy2021an}.}
For ViT, we use 12 transformer layers similar to the one shown in Figure \ref{fig:layer}. The patch size is set to be 16, the hidden dimension for attention is 1024, the hidden dimension for the feedforward network is 4096 with GeLU activation, and the number of heads is 8. The ViT model learns a fixed window update rule, directly mapping 10 input steps to 10 output steps. 

\paragraph{PROSE-FD \cite{liu2024prosefd}.} We directly use the results for PROSE-FD in \cite{liu2024prosefd}, since the same dataset and same context window is used. 

\paragraph{MPP \cite{mccabe2023multiple}.}
In \cite{mccabe2023multiple}, MPP is trained on a different training dataset and different evaluation metrics are used. For a fair comparison, we retrained MPP-B and MPP-L on our training dataset keeping the same model configurations and optimizer hyperparameters. We evaluate MPP-B and MPP-L using the same testing setup and metric, where rollout is used to obtain all 10 output steps.

\paragraph{DPOT \cite{hao2024dpot}.}
In \cite{hao2024dpot}, DPOT is trained on a different training dataset and different evaluation metrics are used. For a fair comparison, we retrained DPOT-M and DPOT-L on our training dataset keeping the same model configurations and optimizer hyperparameters. We evaluate DPOT-M and DPOT-L using the same testing setup and metric, where rollout is used to obtain all 10 output steps.

\subsection{Necessity of Retraining MPP and DPOT for Fairness}\label{sec:mpp_dpot_retrain}
\paragraph{Different Context Window.} MPP is trained using a context window of 16, while DPOT is trained using a context window of 10. As these models learn a fixed window update rule, it is neither possible nor fair to directly use MPP with a context window of 10. This also shows one benefit of BCAT's next frame prediction, i.e., flexible context window. Even though BCAT is trained with 10 input timestamps, it can directly handle longer input sequence lengths such as 16, while DPOT has to discard the first 6 timestamps.   

\paragraph{Different Dataset.} MPP and DPOT are trained on different datasets. MPP was not trained on PDEArena and CFDBench dataset, and DPOT was not trained on PDEBench incompressible NS dataset. Additionally, DPOT did not split a validation dataset, making its training dataset larger than MPP. 

\paragraph{Different Evaluation Metric.} While both MPP and DPOT report their errors using the term relative $L^2$ errors, the quantitative definition may differ, e.g., number of output steps, averaging in space versus space-time, etc. Thus we do not directly compare to the reported numbers in MPP and DPOT. 
~\\

To address this, we retrain MPP and DPOT, using the same context window of 10, the same collection of training dataset, and evaluate them using the same metric. We utilized their code, model configuration, and hyperparameters, and only altered the necessary parts mentioned above. In Table~\ref{tab:pretrain}, we compare the performance of DPOT-M and DPOT-L when we directly evaluate the pretrained checkpoints versus our retrained model. We observe that our retrained version has similar or better performance on all datasets compared to their released pretrained checkpoints.

\begin{table*}[t]
\centering
\caption{\textbf{Evaluating Pretrained Checkpoints vs. Retrained Model.}  * means directly evaluating released pretrained checkpoints. Numbers in () mean that the model is not pretrained on this dataset. The numbers reported are relative $L^2$ errors (\%). The averages are taken with respect to the 6 distinct families listed in the columns of the table.
} 
\label{tab:pretrain}
{
\begin{NiceTabular}{cc|cccccc|c}
\toprule
\multirow{2}{*}{Model} & \multirow{2}{*}{Param} & \multicolumn{3}{c|}{PDEBench}        & \multicolumn{2}{c|}{PDEArena}     & CFDBench & \multirow{2}{*}{Average} \\
                       &                        & SWE & CNS* & \multicolumn{1}{c|}{INS} & NS & \multicolumn{1}{c|}{NS-cond} & -      &                          \\ 
\midrule
DPOT-M     & 122M & 0.54 & {1.01} & 5.20 & {4.92} & {8.55} & 0.64 & {3.47} \\
DPOT-M*     & 122M & 0.32 & {2.82} & (81.57) & {4.49} & {8.40} & 3.20 & - \\
\midrule
DPOT-L     & 523M & 0.15 & {0.89} & 4.08 & {2.21} & {5.29} & 0.34& {2.16} \\
DPOT-L*     & 523M & 0.28 & 2.67 & (66.64) & 4.34 & 7.23 & 3.12 & - \\
\bottomrule
\end{NiceTabular}
}
\end{table*}

\subsection{Remarks on Single-Task Models}\label{sec:single_task_model}

In Table~\ref{tab:main_results}, we compare the zero-shot performance of training a single model for all families of datasets simultaneously. The main focus is to show that these single-task models are not capable of handling more complicated tasks and datasets we are considering using a single pretrained model. For completeness, we also train these single-task models separately on each dataset, and the results are shown in Table~\ref{tab:single_task}. We observe that, the performance does improve for almost all cases (as expected); however, the error's overall order does not change. All single-task models have difficulty predicting the PDEArena dataset, suggesting larger models and better learning algorithms are needed.

\begin{table}
\centering
\caption{\textbf{Single-tasked models trained on all data simultaneously v.s. on each class separately.} * means the results are obtained by training separate models on each dataset. The numbers reported are relative $L^2$ errors (\%).
} 
\label{tab:single_task}
{
\begin{NiceTabular}{lc|cccccc}
\toprule
\multirow{2}{*}{Model} & \multirow{2}{*}{Param} & \multicolumn{3}{c|}{PDEBench}        & \multicolumn{2}{c|}{PDEArena}     & CFDBench  \\
                       &                        & SWE & CNS* & \multicolumn{1}{c|}{INS} & NS & \multicolumn{1}{c|}{NS-cond} & -                             \\ 
\midrule
DeepONet & 3.5M & 3.55 & 7.41 & 64.61 & 35.33 & 51.85 & 12.50 \\
DeepONet* & $6 \times$ 3.5M & 2.86 & 7.31 & 57.61 & 31.00 & 48.34 & 14.08 \\
\midrule
FNO      & 0.6M & 3.71 & 6.31 & 36.84 & 38.67 & 55.63 & 8.52 \\
FNO* & $6 \times$ 0.6M & 1.14 & 5.58 & 21.67 & 26.39 & 44.89 & 3.21 \\
\midrule
UNet     & 5.6M & 0.33 & 3.19 & 3.43 & 12.56  & 16.82 & 0.76 \\
UNet* & $6 \times$ 5.6M & 0.19 & 3.32 & 3.15 & 12.96 & 14.02 & 0.43 \\
\bottomrule
\end{NiceTabular}
}
\end{table}

\section{Additional Results} \label{sec:additional_exps}

Unless stated otherwise, the results in this section use the same held-out test trajectories, autoregressive evaluation protocol, and relative $L^2$ metric as the main comparison in Section~\ref{sec:main_results} (ten output steps, except four for PDEArena NS).

\subsection{Multi-Seed Variability}
Table~\ref{tab:multi-seed} reports the mean and sample standard deviation across five independently seeded BCAT training runs. All runs use the same model, training data, and hyperparameters as the main BCAT experiment; only the random seed changes. The table characterizes the variability of the final test error across the six fluid families. We observe that the standard deviation is small relative to the mean, indicating that BCAT's performance is robust to random initialization.

\begin{table}[h]
\centering
\caption{\textbf{BCAT five seed runs.} The numbers reported are relative $L^2$ errors (\%).
} 
\label{tab:multi-seed}
{
\begin{NiceTabular}{cccccc}
\toprule
\multicolumn{3}{c|}{PDEBench}        & \multicolumn{2}{c|}{PDEArena}     & CFDBench  \\
SWE & CNS* & \multicolumn{1}{c|}{INS} & NS & \multicolumn{1}{c|}{NS-cond} & -                             \\ 
\midrule
$0.12 \pm 0.02$ & $0.40 \pm 0.01$ & $1.27 \pm 0.07$ & $1.65 \pm 0.05$ & $3.22 \pm 0.12$ & $0.44 \pm 0.06$ \\
\bottomrule
\end{NiceTabular}
}
\end{table}

\subsection{Input Noise}

In Table~\ref{tab:input-noise}, we report BCAT's robustness to multiplicative Gaussian noise applied to the physical-field input sequence before the model's standard normalization. For each input value $x$, we use $\tilde{x}=x(1+\sigma\epsilon)$, where $\epsilon\sim\mathcal{N}(0,1)$ is sampled independently over input times, spatial locations, and channels, and $\sigma$ is the noise level in the table. Only the observed input context is corrupted; evaluation uses the clean ground-truth future trajectory. This is a practical stress test for scientific ML settings where observations may be noisy or partially corrupted.

\begin{table*}[t]
\centering
\caption{\textbf{Evaluate BCAT with Noisy Input.} The numbers reported are relative $L^2$ errors (\%). The averages are taken with respect to the 6 distinct families listed in the columns of the table.}
\label{tab:input-noise}
{
\begin{NiceTabular}{c|cccccc|c}
\toprule
\multirow{2}{*}{Noise Level} & \multicolumn{3}{c|}{PDEBench}        & \multicolumn{2}{c|}{PDEArena}     & CFDBench & \multirow{2}{*}{Average} \\
& SWE & CNS* & \multicolumn{1}{c|}{INS} & NS & \multicolumn{1}{c|}{NS-cond} & -      &                          \\
\midrule
0 (no noise) & 0.10 & 0.39 & 1.34 & 1.59 & 3.13 & 0.52 & 1.18 \\
0.01 & 0.17 & 0.52 & 1.43 & 1.70 & 3.17 & 3.49 & 1.75 \\
0.05 & 1.02 & 1.50 & 2.84 & 3.24 & 5.29 & 6.52 & 3.40 \\
0.10 & 2.90 & 2.82 & 5.19 & 5.61 & 12.77 & 11.14 & 6.74 \\
\bottomrule
\end{NiceTabular}
}
\end{table*}

BCAT remains accurate under mild perturbations: at $\sigma=0.01$, the average relative $L^2$ error is 1.75\%, compared with 1.18\% on clean inputs. Errors increase more substantially at higher noise levels, with the largest sensitivity on the CFDBench geometries.

\subsection{Physics Diagnostics}
We report two physics diagnostics on the predicted velocity $(u,v)$ in physical space. Divergence MAE is the spatiotemporal mean of $|\nabla\cdot\hat{\mathbf{u}}|$, computed using a second-order interior finite-difference stencil; we also list the same quantity on the numerical ground truth as a solver floor. Kinetic-energy spectrum error is the relative $L^2$ distance between the radially averaged 2D kinetic-energy spectra of the prediction and the ground truth, averaged over the prediction horizon. We report divergence for the applicable incompressible, obstacle-free datasets and kinetic-energy spectra for velocity-field datasets.

For incompressible flows without immersed obstacles, we compute the interior finite-difference divergence of the predicted velocity and average its magnitude over space and time:
\begin{align}
\operatorname{DivMAE}
&= \frac{1}{T(H-2)(W-2)}
\sum_{t=1}^{T}\sum_{i=2}^{H-1}\sum_{j=2}^{W-1}
\left|
\frac{\hat{u}_{t,i+1,j}-\hat{u}_{t,i-1,j}}{2\Delta x}
+ \frac{\hat{v}_{t,i,j+1}-\hat{v}_{t,i,j-1}}{2\Delta y}
\right|.
\end{align}

For the kinetic-energy diagnostic, let $\mathcal{F}$ denote the 2D DFT and let $\mathbf{q}$ index its frequency grid. We radially average the spectral kinetic energy into integer wavenumber bins:
\begin{align}
\mathcal{E}_t(k)
&= \underset{\lVert\mathbf{q}\rVert\in[k,k+1)}{\operatorname{RadialMean}}
\left[
\frac{1}{2}\left(
\left|\mathcal{F}[u_t](\mathbf{q})\right|^2
+ \left|\mathcal{F}[v_t](\mathbf{q})\right|^2
\right)\right].
\end{align}
We then compare the predicted and ground-truth spectra at every forecast time:
\begin{align}
\operatorname{SpecL2}
&= \frac{1}{T}\sum_{t=1}^{T}
\frac{\left\lVert\mathcal{E}^{\mathrm{pred}}_t-
\mathcal{E}^{\mathrm{gt}}_t\right\rVert_2}
{\left\lVert\mathcal{E}^{\mathrm{gt}}_t\right\rVert_2+10^{-7}}.
\end{align}

\begin{table*}[t]
\centering
\caption{\textbf{Divergence MAE on incompressible flow benchmarks.} Values are averaged over the predicted horizon; the ground-truth residual is included as a solver floor.}
\label{tab:divergence_mae}
{\begin{NiceTabular}{l|ccc}
\toprule
Model & PDEBench INS & PDEArena NS & PDEArena NS-cond \\
\midrule
Ground Truth & 0.0082 & 0.0153 & 0.0303 \\
BCAT & 0.1336 & 0.0157 & 0.0310 \\
DPOT-L & 0.1655 & 0.0150 & 0.0299 \\
MPP-L & 0.3050 & 0.0149 & 0.0320 \\
\bottomrule
\end{NiceTabular}}
\end{table*}

\begin{table*}[t]
\centering
\caption{\textbf{Kinetic-energy spectrum relative $L^2$ error (\%).} The numbers reported are relative $L^2$ distances between the radially averaged 2D kinetic-energy spectra of the prediction and the ground truth, averaged over the predicted time horizon.}
\label{tab:ke_spectrum}
{
\begin{NiceTabular}{c|ccccc}
\toprule
\multirow{2}{*}{Model} & \multicolumn{2}{c|}{PDEBench}        & \multicolumn{2}{c|}{PDEArena}     & CFDBench \\
& CNS* & \multicolumn{1}{c|}{INS} & NS & \multicolumn{1}{c|}{NS-cond} & -      \\
\midrule
BCAT & 1.67 & 1.65 & 1.06 & 1.52 & 0.07 \\
DPOT-L & 1.51 & 1.26 & 0.95 & 1.56 & 0.21 \\
MPP-L & 3.10 & 2.07 & 3.15 & 3.60 & 0.11 \\
\bottomrule
\end{NiceTabular}
}
\end{table*}

On PDEBench INS, all models exceed the numerical ground-truth divergence floor ($8.15\times 10^{-3}$), but BCAT has the lowest residual among the three methods ($0.1336$ versus $0.1655$ for DPOT-L and $0.3050$ for MPP-L). On both PDEArena variants, all three methods are essentially at the solver floor, so divergence is not a useful ranking metric there. The KE-spectrum error is more discriminative: BCAT and DPOT-L are comparable, while MPP-L is consistently worse, especially on PDEArena NS-cond (1.52\% vs. 3.60\%) and PDEBench CNS (1.67\% vs. 3.10\%). Together with the main relative $L^2$ results, this suggests that BCAT's field accuracy is not obtained by sacrificing incompressibility or the kinetic-energy spectrum relative to the strongest baseline.

\subsection{Muon Optimization Diagnostics}\label{sec:muon_diagnostics}
\begin{figure*}[t]
    \centering
    \begin{subfigure}[b]{0.49\linewidth}
        \includegraphics[width=\linewidth]{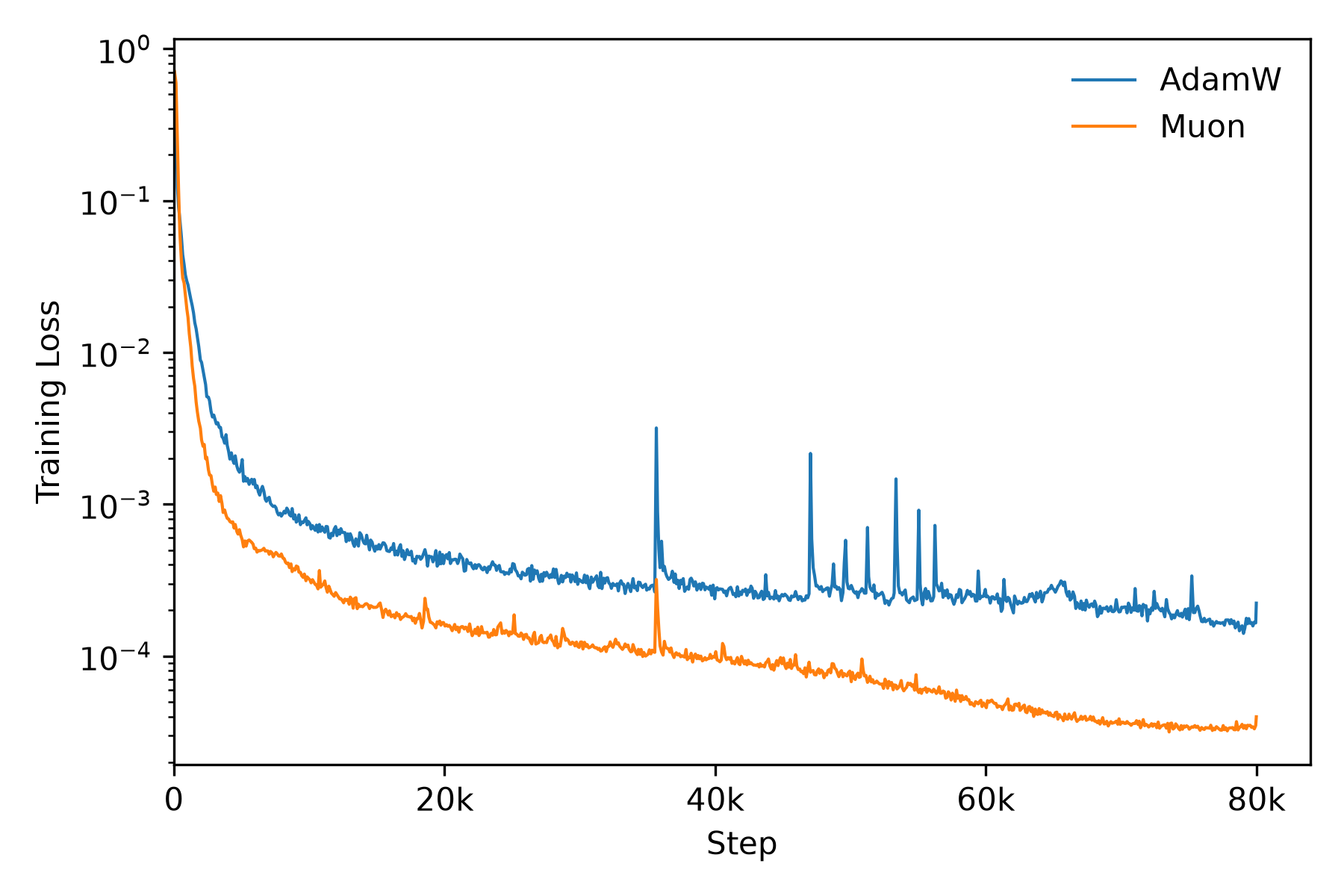}
        \caption{Training loss}
    \end{subfigure}\hfill
    \begin{subfigure}[b]{0.49\linewidth}
        \includegraphics[width=\linewidth]{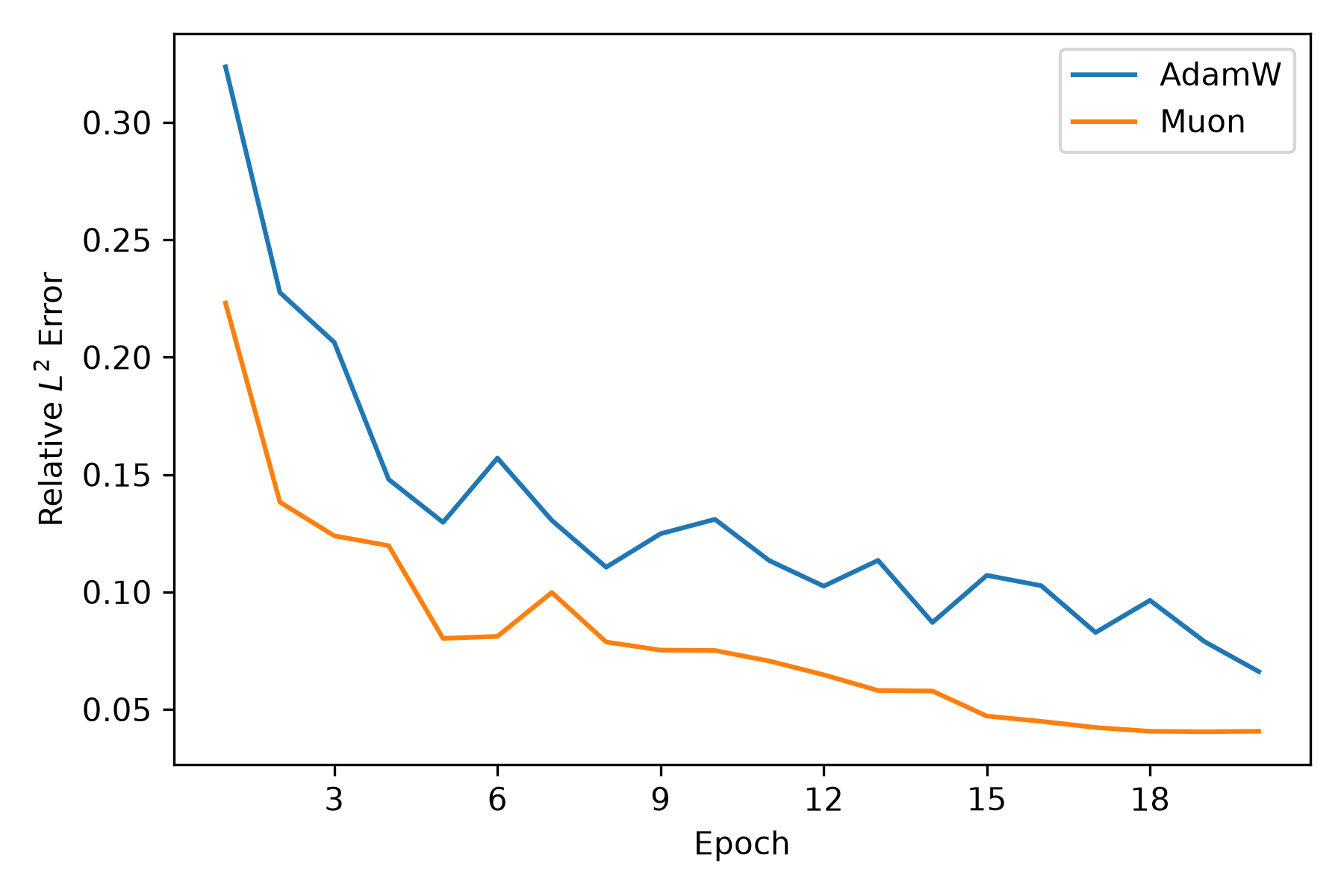}
        \caption{Validation error}
    \end{subfigure}
    \caption{\textbf{Controlled optimizer convergence on PDEArena NS-cond.} BCAT is trained for 20 epochs with AdamW or mixed Muon-AdamW optimization.}
    \Description{Two line plots compare AdamW with mixed Muon-AdamW over a 20-epoch PDEArena conditioned Navier-Stokes training run. Muon has lower training loss and lower validation relative L2 error throughout most of the run.}
    \label{fig:muon_convergence}
\end{figure*}

To examine optimizer behavior in a controlled setting, we train the same BCAT architecture for 20 epochs on PDEArena NS-cond using AdamW or the mixed Muon-AdamW setup. The data split, batch size, and remaining training configuration follow the main experiment. Figure~\ref{fig:muon_convergence} shows lower training loss and validation error for Muon.

\begin{figure}[t]
    \centering
    \includegraphics[width=0.6\linewidth]{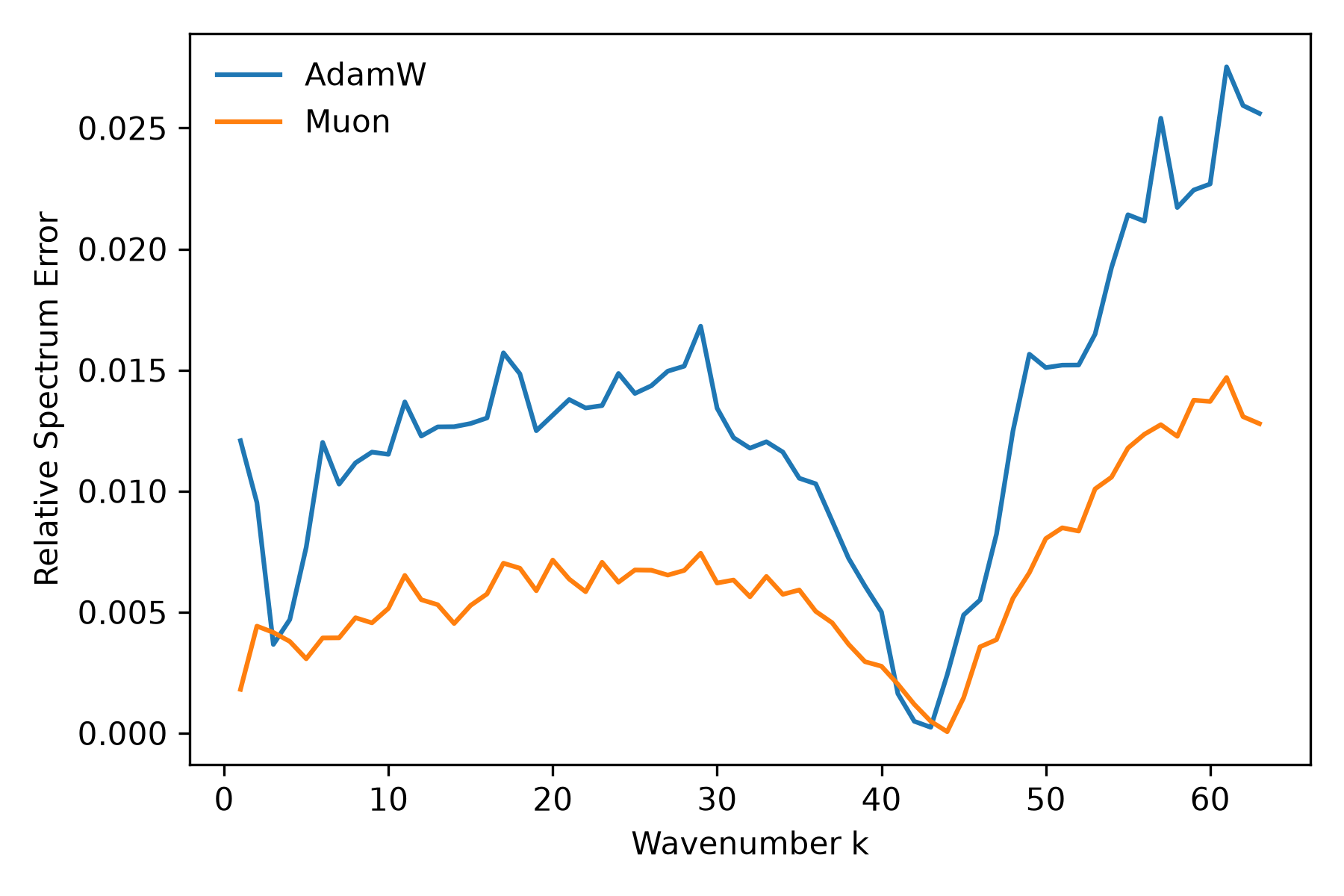}
    \caption{\textbf{Relative speed-spectrum error for AdamW and Muon on PDEArena NS-cond.} Muon has lower error across most resolved wavenumbers, including the high-wavenumber range.}
    \Description{A line plot of relative speed-spectrum error versus radial wavenumber compares AdamW and Muon. Muon is lower at nearly all wavenumbers, with the largest separation at high wavenumber.}
    \label{fig:muon_speed_spectrum}
\end{figure}

Figure~\ref{fig:muon_speed_spectrum} compares a speed-spectrum diagnostic for the AdamW and Muon predictions. Let $s=\sqrt{u^2+v^2}$ be speed on a $H\times W$ grid. We take the orthonormal 2D DFT of $s$, shift the DC component to the center, form the power $P=|\hat{s}|^2$, and average this power within integer radial-wavenumber rings. After averaging over samples and predicted times, we plot for $k=1,\ldots,\min(H,W)/2-1$ the relative error $|P_{\mathrm{pred}}(k)-P_{\mathrm{gt}}(k)|/(P_{\mathrm{gt}}(k)+10^{-10})$. This speed-based diagnostic is distinct from the kinetic-energy spectrum metric in Table~\ref{tab:ke_spectrum}.

\section{Visualizations} \label{sec:more_visual}

\begin{figure*}
\centering
\includegraphics[width=0.75\linewidth]{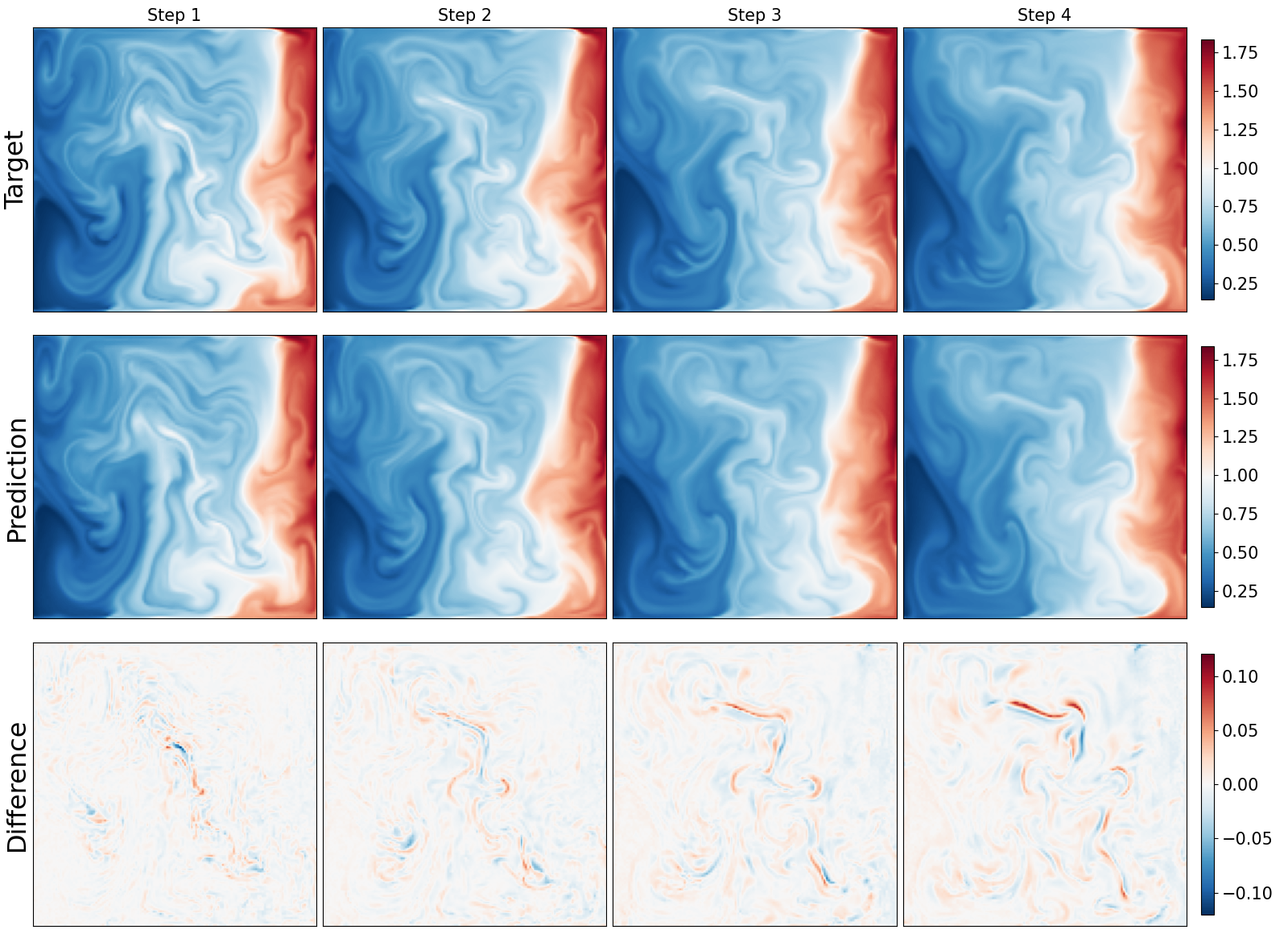}
\caption{\textbf{Example outputs from the BCAT model.} 4 output steps for PDEArena Navier-Stokes dataset. The channel plotted is the particle density in equation \eqref{eq:arena_ns}. Each column represents a different timestamp. For this trajectory, the relative $L^2$ error is 1.93\%.}
\Description{A sequence of particle-density ground-truth and BCAT prediction images for four future PDEArena Navier-Stokes time steps. The predicted structures closely follow the target, with a trajectory-level relative L2 error of 1.93 percent.}
\label{fig:ex_output}
\end{figure*}

\begin{figure}
\centering
\begin{subfigure}[b]{\textwidth}
   \includegraphics[width=1\linewidth]{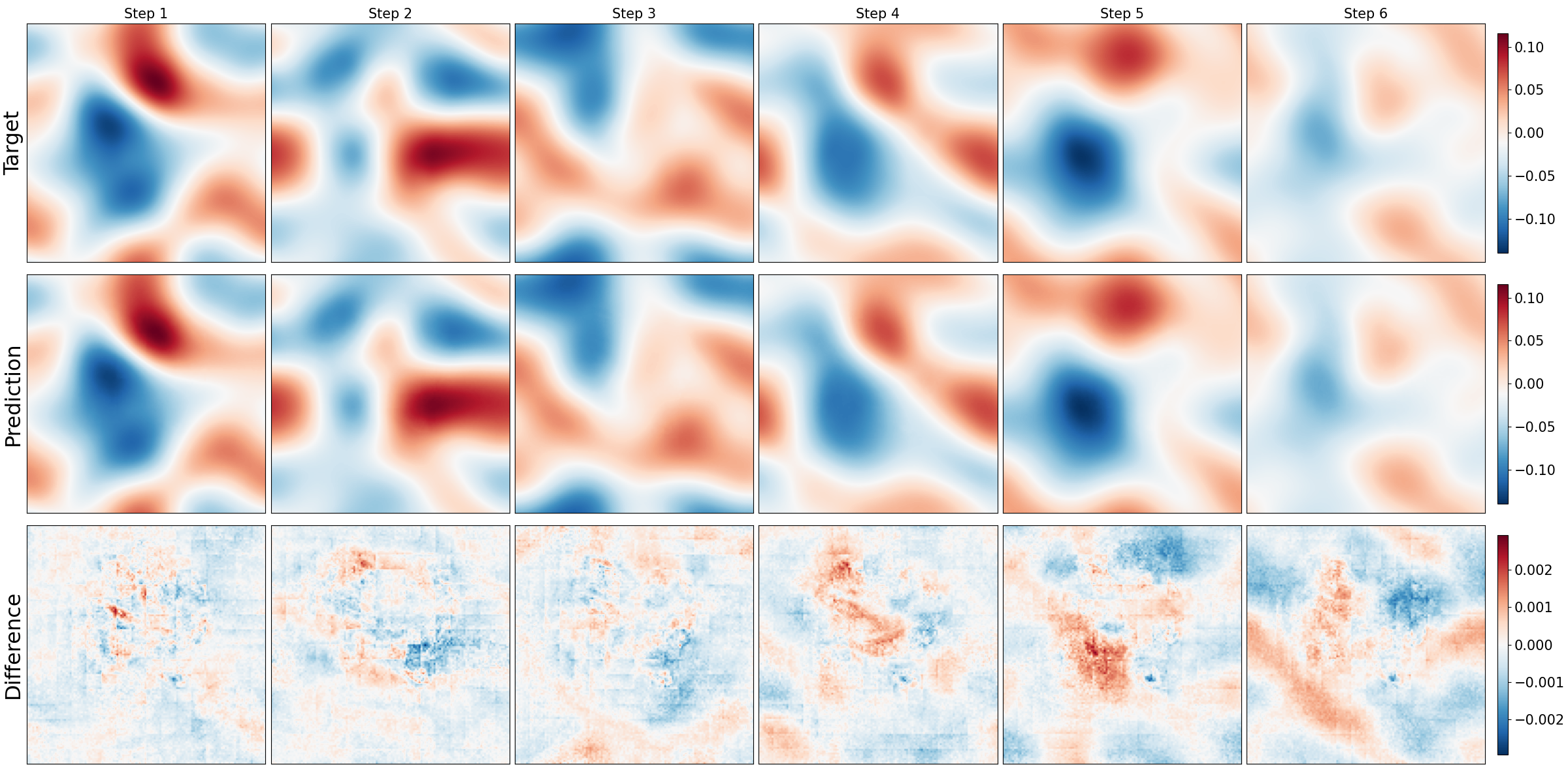}
   \caption{6 output steps for PDEBench Compressible Navier-Stokes dataset. The channel plotted is the $x$-velocity in equation \eqref{eq:cns}. Each column represents a different timestamp. For this trajectory, the relative $L^2$ error is 0.03\%.\\}\label{fig:patch_1}
\end{subfigure}
\begin{subfigure}[b]{\textwidth}
   \includegraphics[width=1\linewidth]{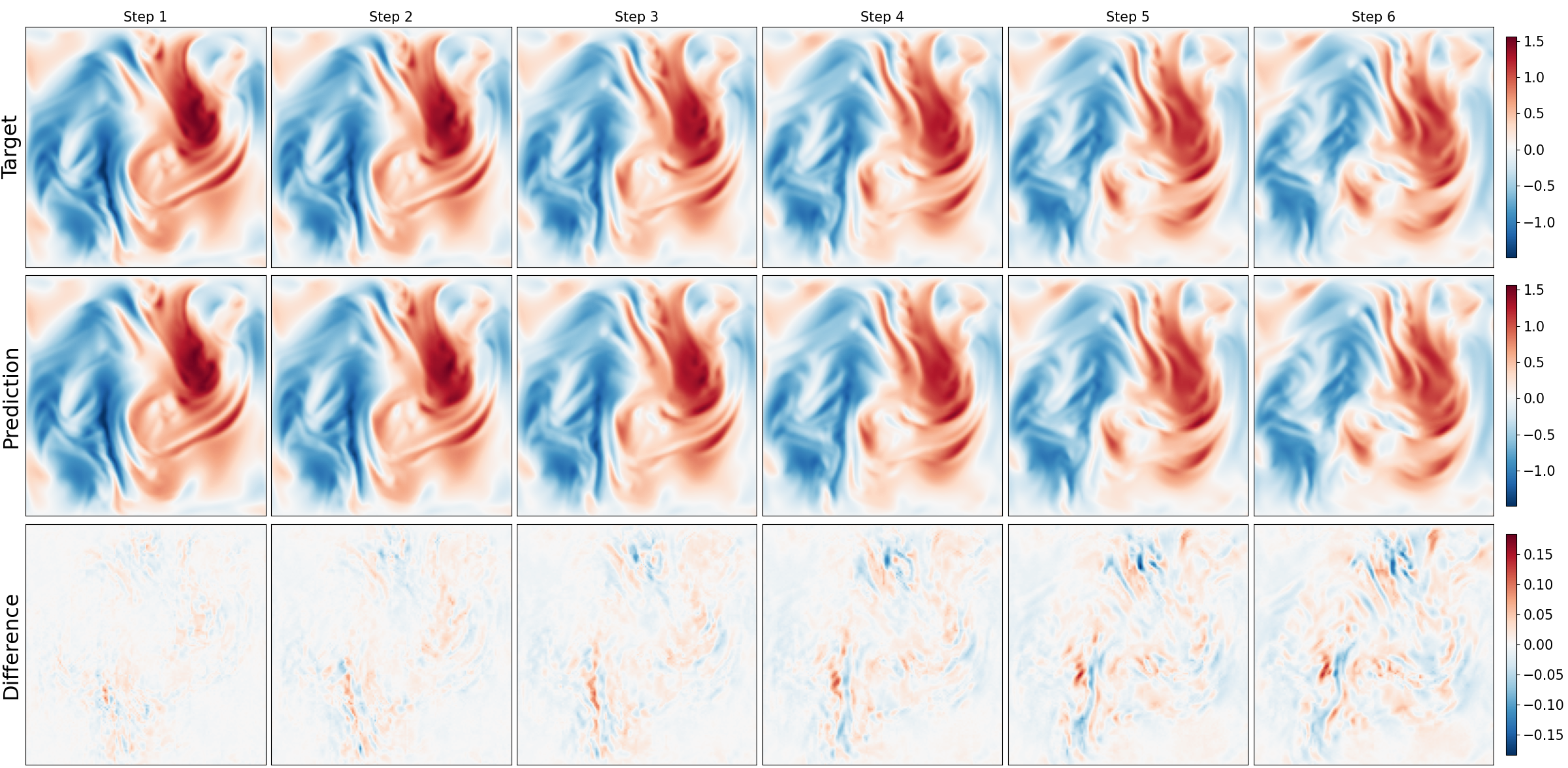}
   \caption{6 output steps for PDEArena Navier-Stokes (conditioned) dataset. The channel plotted is the $x$-velocity in equation \eqref{eq:arena_ns_c}. Each column represents a different timestamp. For this trajectory, the relative $L^2$ error is 4.08\%.}
\end{subfigure}

\caption{\textbf{Two example outputs from the BCAT model.}}
\Description{Two visual examples compare ground-truth and BCAT-predicted velocity fields over six output times: one for compressible Navier-Stokes and one for conditioned Navier-Stokes.}
\label{fig:more_output}
\end{figure}

\begin{figure*}
    \centering
    \includegraphics[width=0.85\linewidth]{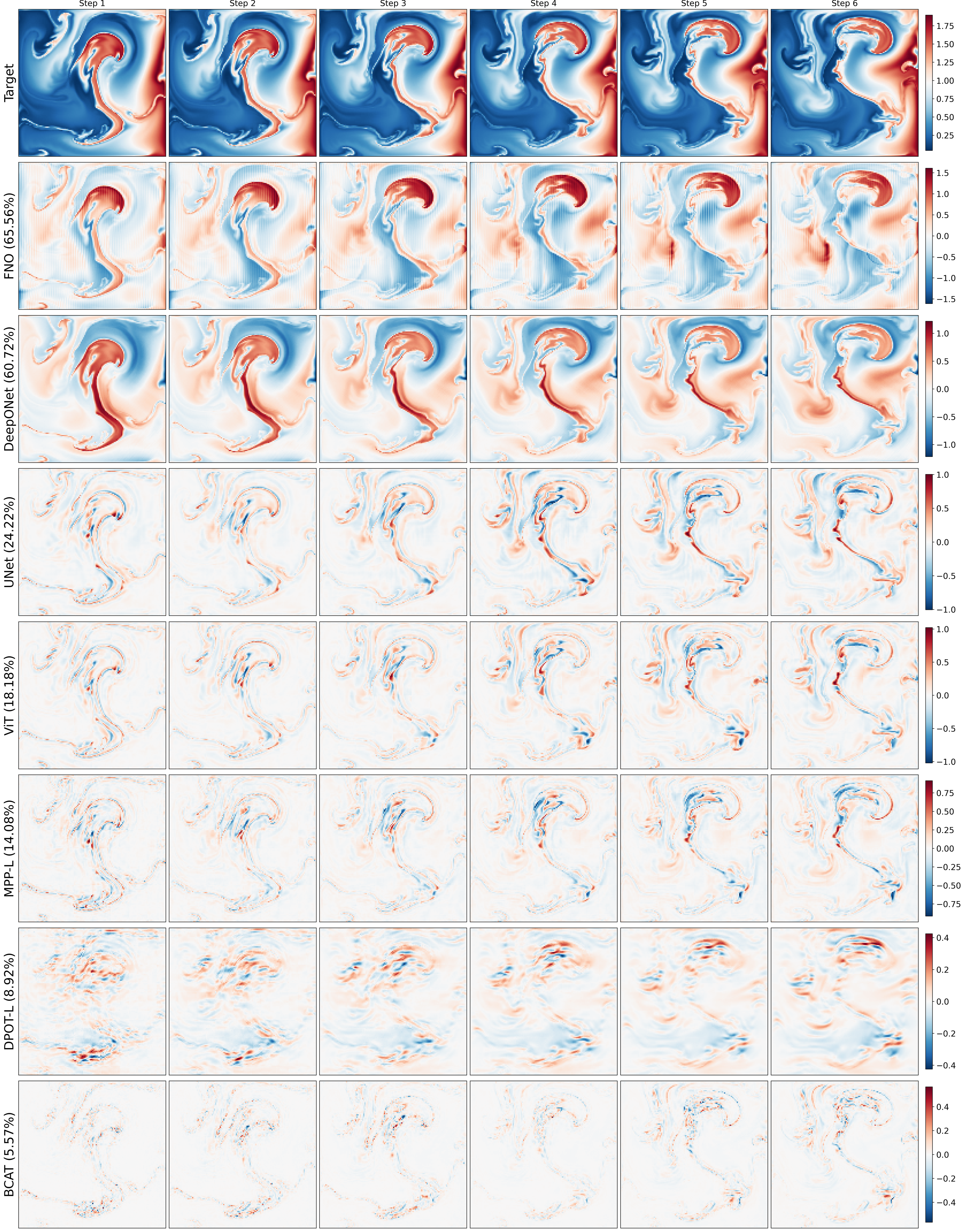}
    \caption{\textbf{Comparing outputs from different models.} The target (first row) is the first 6 output steps from the PDEArena Navier-Stokes (conditioned) dataset (particle density channel). For each model (each row), we display the difference between the target and model output. Relative $L^2$ errors for the full trajectories are listed after the model names.}
    \Description{A grid compares target conditioned Navier-Stokes particle-density frames with error maps from several models over six future steps. Rows correspond to models and the listed trajectory errors summarize each row.}
    \label{fig:compare_output}
\end{figure*}

\begin{figure}[h]
    \centering
    \includegraphics[width=0.8\linewidth]{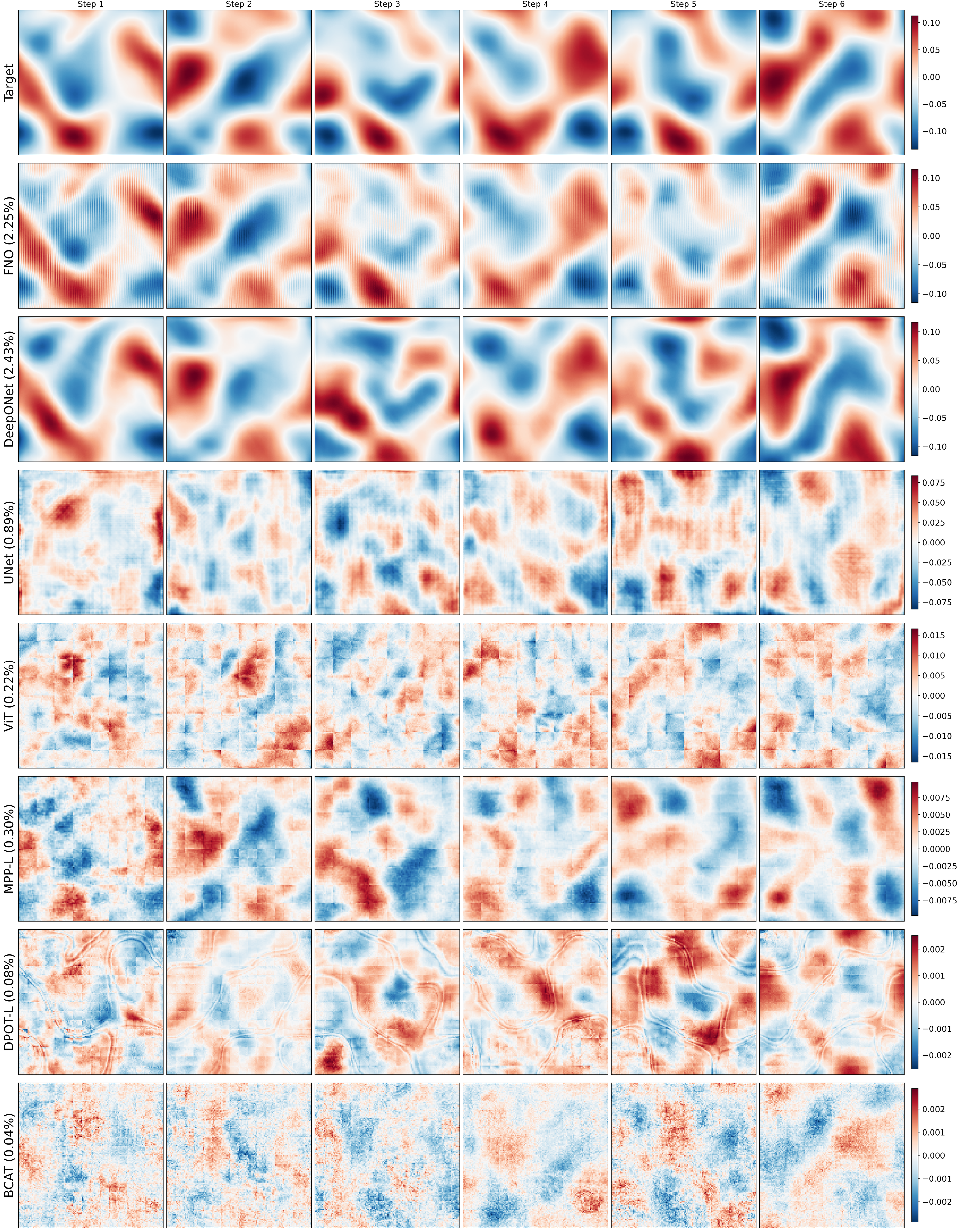}
    \caption{\textbf{Comparing outputs from different models.} The target is the first 6 output steps from the PDEBench Compressible Navier-Stokes dataset ($y$-velocity channel). For each model (each row), we display the difference between target and model output. Relative $L^2$ errors for the full trajectory are listed after the model names.}
    \Description{A grid compares target compressible Navier-Stokes y-velocity frames with error maps from several models over six future steps. Rows correspond to models and the listed trajectory errors summarize each row.}
    \label{fig:compare_output_2}
\end{figure}

In this section, we provide visualizations of the model outputs. Figure \ref{fig:ex_output} and Figure \ref{fig:more_output} contain output visualizations from the BCAT model. In Figure \ref{fig:compare_output} and Figure \ref{fig:compare_output_2}, we compare different model outputs on the same example. While patch effects and patch boundary discontinuities do exist, e.g. Figure~\ref{fig:patch_1} and Figure~\ref{fig:compare_output_2}, which is common for patch-based methods, they are not the main source of error, e.g. high frequency and high contrast regions. To validate this, we compute the average error of interior points (masking out patch boundaries), which is 1.17\% versus the total average error of 1.18\%.  

\clearpage

\end{document}